\documentclass[sigconf]{acmart}

\AtBeginDocument{%
  }

\setcopyright{acmlicensed}
\copyrightyear{2025}
\acmYear{2025}
\acmDOI{XXXXXXX.XXXXXXX}
\acmConference[Conference acronym 'XX]{Make sure to enter the correct
  conference title from your rights confirmation email}{June 03--05,
  2018}{Woodstock, NY}
\acmISBN{978-1-4503-XXXX-X/2018/06}

\acmSubmissionID{1097}

\usepackage{booktabs}
\usepackage{amsmath}
\usepackage{amsthm}

\usepackage{amssymb}
\usepackage{cleveref}
\usepackage{multirow} 
\usepackage{caption}
\usepackage{graphicx}
\usepackage{subcaption} 
\usepackage{mathtools}
\usepackage{algorithmic}
\usepackage{tikz}
\usepackage{comment}
\usepackage{derivative}
\usepackage{makecell}
\usepackage{stfloats}
\usetikzlibrary{arrows.meta, positioning}

\DeclareMathOperator{\generator}{Generator}
\DeclareMathOperator{\VSAL}{VSAL}
\DeclareMathOperator{\thetaGen}{\boldsymbol{\theta_\text{gen}}}
\DeclareMathOperator*{\define}{\coloneqq}
\DeclareMathOperator{\Kamada}{Kamada-Kawai}
\DeclareMathOperator{\Spring}{Spring}
\DeclareMathOperator{\mean}{mean}
\DeclareMathOperator{\M}{\boldsymbol{\mathsf{M}}}
\DeclareMathOperator{\rref}{ref}
\usepackage[ruled]{algorithm2e} 

\copyrightyear{2026}
\acmYear{2026}
\setcopyright{cc}
\setcctype{by}
\acmConference[WWW '26]{Proceedings of the ACM Web Conference 2026}{April 13--17, 2026}{Dubai, United Arab Emirates}
\acmBooktitle{Proceedings of the ACM Web Conference 2026 (WWW '26), April 13--17, 2026, Dubai, United Arab Emirates}
\acmISBN{979-8-4007-2307-0/2026/04}
\acmDOI{10.1145/3774904.3792224}

\begin{document}

\title{VSAL: A Vision Solver with Adaptive Layouts for Graph Property Detection}

\author{Jiahao Xie}
\email{jiahaox@udel.edu}
\orcid{0009-0002-8716-6566}
\affiliation{%
  \institution{University of Delaware}
  \city{Newark}
  \state{Delaware}
  \country{USA}
}

\author{Guangmo Tong}
\email{amotong@udel.edu}
\orcid{0000-0003-3247-4019}
\affiliation{%
  \institution{University of Delaware}
  \city{Newark}
  \state{Delaware}
  \country{USA}
}

\renewcommand{\shortauthors}{Jiahao Xie and Guangmo Tong}

\begin{abstract}
Graph property detection aims to determine whether a graph exhibits certain structural properties, such as being Hamiltonian. Recently, learning-based approaches have shown great promise by
leveraging data-driven models to detect graph properties efficiently. In particular, vision-based methods offer a visually intuitive solution by processing the visualizations of graphs.
However, existing vision-based methods rely on fixed visual graph layouts, and therefore, the expressiveness of their pipeline is restricted. 
To overcome this limitation, we propose VSAL, a vision-based framework that incorporates an adaptive layout generator capable of dynamically producing informative graph visualizations tailored to individual instances, thereby improving graph property detection. 
Extensive experiments demonstrate that VSAL outperforms state-of-the-art vision-based methods on various tasks such as Hamiltonian cycle, planarity, claw-freeness, and tree detection.
\end{abstract}




\maketitle
\newcommand\webconfavailabilityurl{https://doi.org/10.5281/zenodo.18332925}
\ifdefempty{\webconfavailabilityurl}{}{
\begingroup\small\noindent\raggedright\textbf{Resource Availability:}\\
The source code and data of this paper have been released at \url{\webconfavailabilityurl} and \url{https://github.com/Jiahao-Xie-86/VSAL}.
\endgroup
}

\section{Introduction}
Graphs are a fundamental data structure for modeling the web, representing various structures such as hyperlink networks \cite{broder2000graph}, social networks \cite{wang2024query, wang2025data}, and clickstream data \cite{meusel2014graph}. Detecting structural properties of these web graphs, such as Hamiltonicity \cite{gould2014recent}, planarity \cite{hopcroft1974efficient}, and connectivity \cite{carriere1997webquery}, is significant for a wide range of web applications, including community detection \cite{leskovec2010empirical}, network visualization \cite{shneiderman2006network}, hyperlink prediction \cite{chen2023survey}, and anomaly detection~\cite{jindal2007review}. However, traditional algorithmic methods for graph property detection often require special designs for individual properties and also struggle with scalability \cite{bianchi2009survey}.
More recently, statistical learning methods 
have emerged as data-driven alternatives for handling such tasks,
providing greater flexibility and efficiency \cite{wu2020comprehensive, xie2025advances}.

\begin{figure}[t!]
    \centering
\subfloat[Simple Hamiltonian]{\label{fig:VSGL_simple_ham}\includegraphics[width=0.2\textwidth]{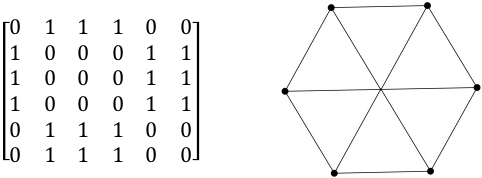}}\hspace{9mm}
\subfloat[Simple Planar]{\label{fig:VSGL_simple_planar}\includegraphics[width=0.2\textwidth]{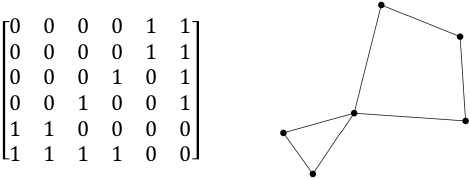}}

\subfloat[Simple Claw-free]{\label{fig:VSGL_simple_claw}\includegraphics[width=0.2\textwidth]{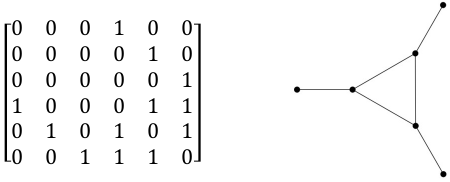}}\hspace{9mm}
\subfloat[Simple Tree]{\label{fig:VSGL_simple_tree}\includegraphics[width=0.2\textwidth]{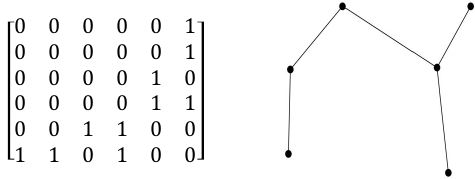}}

\subfloat[Hamiltonian]{\label{fig:VSGL_complex_ham}\includegraphics[width=0.15\textwidth]{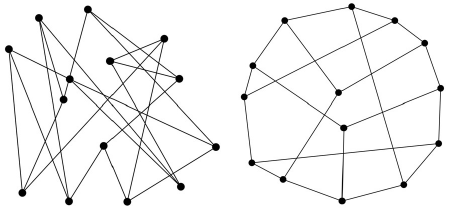}}\hspace{1.5mm}
\subfloat[Planar]{\label{fig:VSGL_complex_planar}\includegraphics[width=0.15\textwidth]{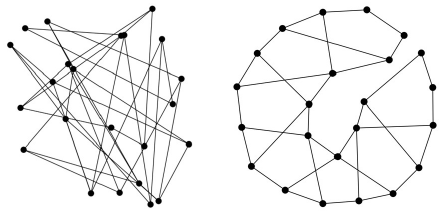}}\hspace{1.5mm}
\subfloat[Claw-free]{\label{fig:VSGL_complex_claw}\includegraphics[width=0.15\textwidth]{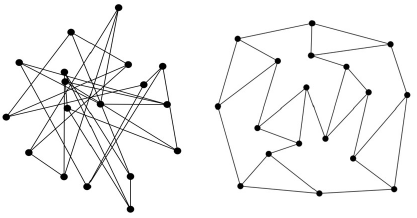}}
    \caption{Each pair of (a)-(d) shows a simple graph with a certain property, including the adjacency matrix (left) and a well-designed layout generated by VSAL (right). Each pair of (e)-(g) shows a complex graph with a random layout (left) and a well-designed layout generated by VSAL (right). 
    }
    \Description{The figures illustrating the intuitive advantage of vision-based methods over matrix-based methods.}
    \label{fig:VSGL_layout_framework}
\end{figure} 

Despite their advantages, existing learning methods primarily encode graph information through adjacency matrices, making them less effective for tasks involving spatial and visual interpretation \cite{schuetz2022combinatorial}, such as identifying geometric symmetries~\cite{dwivedi2023benchmarking}.
In this regard, vision-based methods offer a promising alternative~\cite{samizadeh2023vn,ling2020solving,ling2023deep}. One particular pipeline of such methods is VN-Solver \cite{samizadeh2023vn} consisting of  
three conceptual steps: a) transforming graphs into 2D layouts, b) visualizing layouts as images, and c) detecting graph properties by image classifiers.
Such a pipeline explores an interesting concept: can we detect graph properties by taking a look at graph visualizations? Intuitively, many graph properties can be much more evident from visualizations than from matrices; for example, the graph properties in Figs.~\ref{fig:VSGL_simple_ham}-\ref{fig:VSGL_simple_tree} can be easily confirmed from well-designed layouts, but it is arguably less straightforward for humans to do the same by scrutinizing the adjacency matrices. As for larger and more complex graphs, visual patterns may become challenging to interpret manually (e.g., Figs.~\ref{fig:VSGL_complex_ham}-\ref{fig:VSGL_complex_claw}), but recent advances in image classification provide effective tools for such purposes, such as ResNet~\cite{he2016deep} and Vision Transformer (ViT)~\cite{dosovitskiy2021an}. Indeed, it has been shown that such a framework is not only feasible but also comparable to the state-of-the-art matrix-based methods for the Hamiltonian cycle problem~\cite{samizadeh2023vn}.

\textbf{Contribution}. Although VN-solver has achieved promising performance, it relies on fixed layouts that are not adaptable to the underlying data distribution. In this light, we propose \textbf{V}ision neural \textbf{S}olver with \textbf{A}daptive \textbf{L}ayouts (\textbf{VSAL}), a new framework aiming to create layouts that can mimic principled designs while being learnable to capture instance-specific properties based on classification feedback.
To realize such an idea, we leverage the idea of generative adversarial networks \cite{goodfellow2014generative} to generate dynamic layouts, and design a differentiable visualization module to enable end-to-end training. 
To our delight, 
VSAL can learn to generate layouts that can not only visually clarify graph structural properties 
(Figs.~\ref{fig:VSGL_complex_ham}-\ref{fig:VSGL_complex_claw}), but also enable significantly better detection performance than VN-Solver on multiple graph property detection tasks (as shown in Sec.~\ref{sec:experiment}).
Furthermore, VSAL functions as intended; for example, it can effectively optimize various principled graph layouts, and the generated layouts exhibit more clarity (e.g., reducing edge crossings) and can capture key features for better detection (e.g., identifying isolated and one-degree nodes for Hamiltonian cycle detection).


\section{Related Work}
\label{sec:related}
\textbf{Matrix-based learning methods}. Existing statistical learning methods for graph property detection are primarily matrix-based, as summarized in recent surveys 
\cite{ma2021comprehensive,li2024comprehensive}. 
Graphormer \cite{ying2021transformers} adapts the Transformer \cite{vaswani2017attention} to graphs by incorporating three structural encodings: centrality (capturing node importance via degree measures), spatial (encoding pairwise shortest-path distances), and edge (integrating edge features into attention). These enhancements enable Graphormer to capture global dependencies and achieve state-of-the-art performance on graph property detection tasks \cite{koolattention,dwivedi2023benchmarking}. 
Building on this architecture, Graphormer-GD \cite{zhangrethinking} introduces Gaussian kernel-based distance encoding to better model structural similarities and improve generalization ability. 
EquiformerV2~\cite{liaoequiformerv2} extends the Transformer with SE(3)-equivariance, allowing it to exploit geometric symmetries and directional information, which is especially beneficial for spatially structured graphs. 
More recently, GraphsGPT \cite{gao2024graph} introduces a pure Transformer architecture that encodes graphs as sequences of learned Graph Words via a Graph2Seq encoder and reconstructs them through an edge-centric autoregressive decoder, yielding expressive representations for downstream tasks.
While effective, these matrix-based methods are less capable of capturing visual patterns, as the adjacency matrices primarily represent relational node interactions 
\cite{cappart2023combinatorial}.


\textbf{Vision-based learning methods}. 
There have been several attempts to integrate vision techniques into optimization tasks involving graphs. For example, Ling \textit{et al.} \cite{ling2023deep,ling2020solving} employ CNNs to directly process raw graph visualizations without designing specific layouts, and   
Graikos \textit{et al.} \cite{graikos2022diffusion} explore diffusion models for iterative graph pattern generation. However, these methods cannot be directly applied to graph property detection tasks. VN-Solver~\cite{samizadeh2023vn} is the first vision-based method that utilizes visualized graph layouts for graph property detection.
By employing ResNet-50 to process the visualizations of fixed circular and spiral graph layouts, VN-Solver tackles the Hamiltonian cycle problem and achieves encouraging performance. However, its reliance on fixed layouts limits its expressiveness and flexibility in capturing complex graph structures. 
In contrast, VSAL overcomes these limitations by leveraging a Wasserstein adversarial generation process~\cite{arjovsky2017wasserstein} to adaptively generate graph layouts under the guidance of a classifier, thereby improving detection accuracy and efficiency.

\textbf{Graph layout generation strategies}. Traditional graph layout algorithms, such as force-directed methods \cite{kamada1989algorithm}, visualize graphs by optimizing certain aesthetic criteria, such as minimizing edge crossings. However, these methods require manual parameter tuning and lack generalization ability across diverse graphs. Recent work has introduced generative models for graph layout generation. For instance, Kwon \textit{et al.} \cite{kwon2019deep} develop a variational autoencoder that learns layout distributions from examples, enabling smooth transitions between different layout styles. Similarly, Wang \textit{et al.}~\cite{wang2023smartgd} propose a generative adversarial framework that optimizes layouts for diverse aesthetic goals such as stress minimization and crossing angle maximization. While these approaches focus on producing aesthetically pleasing layouts, our work takes a different direction by generating layouts explicitly optimized to improve graph property detection accuracy. By conditioning layout generation on classification feedback, VSAL creates layouts that make structural properties more visually distinguishable for the classifier.

\section{Preliminary}
\label{sec:prelimary}
Graph property detection is, in essence, a binary classification task.
Let $\mathcal{G}$ denote the space of all undirected graphs and $\mathcal{Y} = \{0, 1\}$ represent the set of binary labels, where $y=1$ (resp., $y=0$) indicates the presence (resp., absence) of the target property. A graph $G\in\mathcal{G}$ is given by $G=(V,E,\mathbf{A})$, where $V$ is the node set, $E$ is the edge set, and $\mathbf{A} \in \{0,1\}^{|V|\times |V|}$ is the adjacency matrix.
Taking a perspective of statistical learning, each graph property detection task is associated with an unknown underlying distribution $\mathcal{D}$ over $\mathcal{G} \times \mathcal{Y}$. Given a training set $D = \big\{(G_i, y_i)\big\}_{i=1}^n \sim \mathcal{D}$,
the goal is to design a hypothesis space $\mathcal{H} \subseteq \{h: \mathcal{G} \rightarrow \mathcal{Y}\}$ and learn a function $h \in \mathcal{H}$ that can minimize the true loss:
\begin{align*}
\mathcal{L}_{\mathcal{D}}(h) \define \mathbb{E}_{(G,y)\sim \mathcal{D}}\Big[\ell\big(h(G), y\big)\Big],
\end{align*}
where $\ell: \mathcal{Y} \times \mathcal{Y} \rightarrow \mathbb{R}^+$ measures the difference between the prediction $h(G)$ and the ground truth $y$.

\section{Methodology}
\label{sec:method}
The pipeline of vision-based methods for graph property detection consists of three abstract steps. 
\begin{itemize}
    \item \textbf{Layout generation}. Given the input graph $G=(V, E, \mathbf{A})\in \mathcal{G}$, the generator produces a 2D layout based on the structural information of the graph. 
    \begin{align*}    \generator:\mathcal{G}\rightarrow\mathbb{R}^{|V|\times 2}.
    \end{align*}
    \item \textbf{Layout visualization}. The generated 2D graph layout is converted into an RGB image with a resolution of $H \times W$. 
    \begin{align*}
        \mathrm{Visualization}:\mathbb{R}^{|V|\times 2}\rightarrow \mathbb{R}^{H\times W \times 3}.
    \end{align*}
    \item \textbf{Image classification}. The RGB image is subsequently processed by an image classifier to detect the graph properties, outputting a binary label.
    \begin{align*}
        \mathrm{Classifier}:\mathbb{R}^{H\times W \times 3}\rightarrow \{0,1\}.
    \end{align*}
\end{itemize}
Our \textbf{V}ision neural \textbf{S}olver with \textbf{A}daptive \textbf{L}ayouts (\textbf{VSAL}) framework follows such a pipeline but with novel designs for layout generation and visualization. In what follows, we introduce each component of VSAL (Sec.~\ref{subsec:framework}), followed by its training methods (Sec.~\ref{subsec:training}).

\subsection{VSAL Framework}
\label{subsec:framework}

\subsubsection{Layout Generation} 
In contrast to the current methods that utilize deterministic layouts, VSAL seeks to create flexible layouts that can be tuned to capture useful structural features for accurate detection. To this end, we design a class $\mathcal{G}_{\text{gen}}$ of random generators that produce graph layouts conditioned on a latent variable $\mathbf{z} \in \mathcal{Z}=\mathbb{R}^{d_{\mathcal{Z}}}$ of dimension $d_{\mathcal{Z}} \in \mathbb{Z}^{+}$, where each generator $g_{\boldsymbol{\theta}_\text{gen}}$ is parameterized by a nominal set $\boldsymbol{\theta}_\text{gen}$ of parameters: 
\begin{align}
\label{eq:generators}
    \mathcal{G}_{\text{gen}} \subseteq \Big\{g_{\boldsymbol{\theta}_\text{gen}}(G,\mathbf{z}):\mathcal{G}\times\mathcal{Z}\rightarrow\mathbb{R}^{|V| \times 2}\Big\}.
\end{align} 
In particular, given an input graph $G=(V, E, \mathbf{A})$, its layout is generated by the following four steps: a) graph encoding, b) noise encoding, c) concatenation, and d) coordinate generation.

\textbf{a) Graph encoding}. The input adjacency matrix $\mathbf{A}\in\mathbb{R}^{|V|\times |V|}$ is processed through a graph encoder 
to extract structural features:
\begin{align*}
    H_G &= \text{GraphEncoder}_{\theta_\text{gen}^{1}}(\mathbf{A}) \in \mathbb{R}^{|V| \times d_g},
\end{align*}
where $d_g \in \mathbb{Z}^+$ denotes the dimension of the hidden space and $\theta_\text{gen}^{1}$ represents learnable parameters. We can adopt various graph representation models for the graph encoder, such as DenseGCN~\cite{guo-etal-2019-densely} and Graphormer \cite{ying2021transformers}.


\textbf{b) Noise encoding}. Simultaneously, a latent noise vector $\mathbf{z}\in\mathcal{Z}$ is sampled from a Gaussian distribution $P_{\mathcal{Z}}=\mathcal{N}(0,\sigma^2\mathbf{I})$, where $\sigma^2$ is the variance and $\mathbf{I}$ represents the identity matrix. The noise vector $\mathbf{z}$ is then encoded into the same hidden space via a multilayer perceptron $\text{MLP}_{\theta_\text{gen}^{2}}$ with learnable parameters $\theta_\text{gen}^{2}$, 
introducing stochasticity to help generate diverse graph layouts \cite{goodfellow2014generative}.
\begin{align}
    H_z = \text{MLP}_{\theta_\text{gen}^{2}}(\mathbf{z}) \in \mathbb{R}^{|V| \times d_g}.
    \nonumber
\end{align}


\textbf{c) Concatenation}. The graph embedding $H_G$ is concatenated with the noise encoding $H_z$ to form the node representation
\begin{align*}
    \mathbf{H}_{\text{cond}} = \text{concat}\big(H_G, H_z\big) \in \mathbb{R}^{|V| \times 2d_g},
\end{align*}
where the dimension of each node representation is $2d_g$.

\textbf{d) Coordinates generation}. Finally, $\mathbf{H}_{\text{cond}}$ is processed through a multilayer perceptron $\text{MLP}_{\theta_\text{gen}^{3}}$ parameterized by $\theta_\text{gen}^{3}$ to compute the 2D coordinates for each node $i$:
\begin{align*}
    \big(\hat{x}_i^1,\hat{x}_i^2\big) = \text{MLP}_{\theta_\text{gen}^{3}}\big({\mathbf{H}_{\text{cond}}(i)}\big) \in \mathbb{R}^{2}.
\end{align*}

In summary, the generator maps the input graph and the noise vector to the resulting graph layout $g_{\boldsymbol{\theta_\text{gen}}}(G,\mathbf{z})$, i.e., 
\begin{align}
\label{eq: generator}
    g_{\boldsymbol{\theta_\text{gen}}}(G,\mathbf{z}) = \Big[\big(\hat{x}_1^1, \hat{x}_1^2\big), \dots, \big(\hat{x}_{|V|}^1, \hat{x}_{|V|}^2\big)\Big] \in \mathbb{R}^{|V| \times 2},
\end{align}
and the learnable parameters of the generator are collected as
\begin{align*}
    \boldsymbol{\theta_\text{gen}}=\Big\{\theta_\text{gen}^{1},\theta_\text{gen}^{2},\theta_\text{gen}^{3}\Big\}.
\end{align*}

\begin{figure}[t!]
    \centering  
    \subfloat[]{\label{fig:visualization_a}\includegraphics[width=0.10\textwidth]{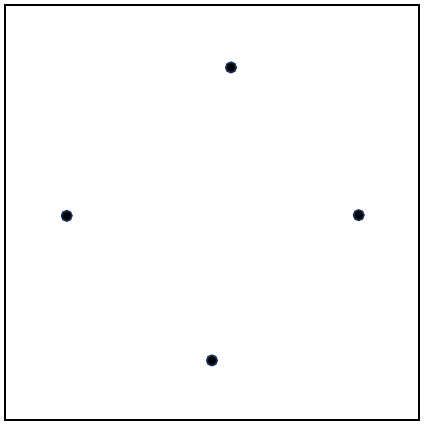}}\hspace{1.4mm}
    \subfloat[]{\label{fig:visualization_b}\includegraphics[width=0.10\textwidth]{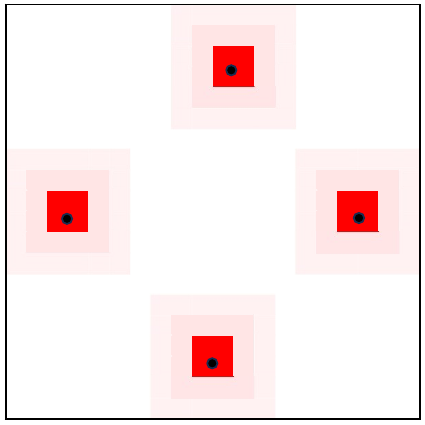}}\hspace{1.4mm}
    \subfloat[]{\label{fig:visualization_c}\includegraphics[width=0.10\textwidth]{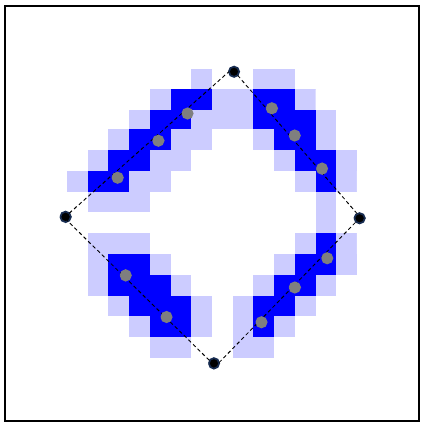}}\hspace{1.4mm}
    \subfloat[]{\label{fig:visualization_d}\includegraphics[width=0.10\textwidth]{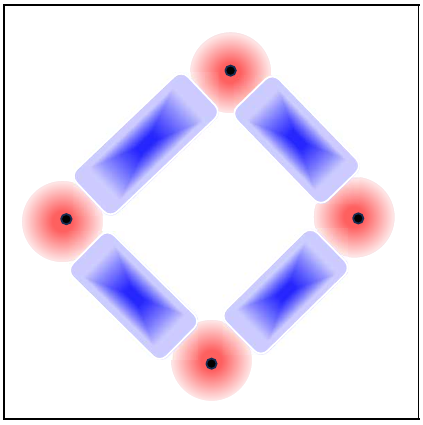}}
    \Description{This figure shows the visualization process of our module.}
    \caption{Layout visualization process: (a) normalized coordinates, (b) node rendering, (c) edge rendering, and (d) Gaussian smoothing.}
    \label{fig:visualization_process}
\end{figure} 

\subsubsection{Layout Visualization}
\label{sec:visualization}
In converting the generated layout $g_{\boldsymbol{\theta_\text{gen}}}(G,\mathbf{z})$ into an RGB image, the coordinates  $\big(\hat{x}_i^1, \hat{x}_i^2\big)$ of each node are first normalized to fit within the image resolution of $[0, H] \times [0, W]$ through
\begin{align*}
\big(\tilde{x}_i^1,\tilde{x}_i^2\big)=\left( \frac{\hat{x}_i^1 - \hat{x}^1_{\min}}{\hat{x}^1_{\max} - \hat{x}^1_{\min}}\cdot H, \frac{\hat{x}_i^2 - \hat{x}^2_{\min}}{\hat{x}^2_{\max} - \hat{x}^2_{\min}}\cdot W \right ), 
\end{align*}
where $\hat{x}^1_{\min}$, $\hat{x}^1_{\max}$, $\hat{x}^2_{\min}$, and $\hat{x}^2_{\max}$ are the minimum and maximum values across all node coordinates in $g_{\boldsymbol{\theta_\text{gen}}}(G,\mathbf{z})$. After that, an image tensor $\boldsymbol{\mathsf{M}} \in \mathbb{R}^{H \times W \times 3}$ 
is initialized with all pixels set to $(255,255,255)$. To ensure the differentiability of the visualization process, we circumvent the need for manual pixel coloring by using Gaussian falloff \cite{drori2003fragment} to achieve smooth rendering. These visualization steps are illustrated in Fig.~\ref{fig:visualization_process}.

\textbf{a) Node rendering}. In visualizing the nodes, for each pixel $\M(p,q,:)$, its closeness to the nodes is measured by
\begin{align*}
    \alpha_{V}(p, q) = 1 - \prod_{i=1}^{|V|} \Bigg(1 - \exp\bigg(-\frac{d_i^2(p, q)}{2r^2}\bigg)\Bigg),
\end{align*}
where $d_i(p, q)$ is the Euclidean distance between $(p,q)$ and $\big(\tilde{x}_i^1,\tilde{x}_i^2\big)$, and $r \in \mathbb{R}^+$ controls the degree of node influence. The RGB channel of $\boldsymbol{\mathsf{M}}$ is then updated by
\begin{align*}
    \boldsymbol{\mathsf{M}}(p, q,:) = \big(1 - \alpha_{V}(p, q)\big)\cdot \boldsymbol{\mathsf{M}}(p, q,:) + \alpha_{V}(p, q)\cdot (255, 0, 0).
\end{align*}  
Intuitively, the pixels closer to the node coordinates will be assigned with $(255,0,0)$ of less transparency---Fig.~\ref{fig:visualization_b}. 

\textbf{b) Edge rendering}. Each edge $(i, j) \in E$ is associated with a set $ L(i, j)$ of $N \in \mathbb{Z}^+$ points uniformly sampled along the line segment between nodes $(\tilde{x}_i^1, \tilde{x}_i^2)$ and $(\tilde{x}_j^1, \tilde{x}_j^2)$: 
\begin{align*}
    L(i, j) \define \bigg\{\frac{N+1-k}{N+1} \big(\tilde{x}_i^1, \tilde{x}_i^2\big) +  \frac{k}{N+1} \big(\tilde{x}_j^1, \tilde{x}_j^2\big)\bigg\}_{k=1}^{N}.
\end{align*}
Similarly, for each pixel $\M(p,q,:)$, its closeness to the edges is measured by
\begin{align*}
    \alpha_{E}(p, q) = 1 - \prod_{(i, j) \in E} \Bigg(1 - \exp\bigg(-\frac{d_{ij}^2(p, q)}{2\delta^2}\bigg)\Bigg),
\end{align*}
where $\delta\in \mathbb{R}^+$ controls the degree of edge influence, and $d_{ij}(p, q)$ is a smooth approximation of the Euclidean distance between $(p,q)$ and $L(i,j)$, which is given by
\begin{align*}
    d_{ij}(p, q) = \frac{\sum_{(x_1, x_2) \in L(i, j)} \exp\Big(-\beta \cdot d_{x_1,x_2}^{p, q}\Big) \cdot d_{x_1, x_2}^{p, q}}{\sum_{(x_1, x_2) \in L(i, j)} \exp\Big(-\beta \cdot d_{x_1, x_2}^{p, q}\Big)}, 
\end{align*}
where $\beta \in \mathbb{R}^+$ controls the sharpness of approximation and $d_{x_1,x_2}^{p, q}$ is the Euclidean distance between $(p,q)$ and  $(x_1, x_2) \in L(i, j)$. The RGB channel of $\boldsymbol{\mathsf{M}}$ is updated by
\begin{align*}
    \boldsymbol{\mathsf{M}}(p, q,:) = \big(1 - \alpha_{E}(p, q)\big)\cdot \boldsymbol{\mathsf{M}}(p, q,:) + \alpha_{E}(p, q) \cdot (0, 0, 255),
\end{align*}
with the implication that pixels closer to the edges will have a stronger color in $(0, 0, 255)$---Fig.~\ref{fig:visualization_c}.

\textbf{c) Gaussian smoothing}. Finally, we apply Gaussian smoothing to reduce pixelation and ensure smooth color transitions \cite{simard2003best}, which is implemented via
\begin{align}
    \boldsymbol{\mathsf{M}}(p, q,:) = \sum_{u=-K}^K \sum_{v=-K}^K \text{Gau}(u, v) \cdot \boldsymbol{\mathsf{M}}(p+u, q+v,:),
    \label{eq:gaussian_smoothing}
\end{align}
where $\text{Gau}(u, v) = \frac{1}{2\pi\sigma^2} \exp\left(-\frac{u^2 + v^2}{2\sigma^2}\right)$ denotes the Gaussian kernel with radius $K\in\mathbb{Z}^+$, and $\sigma\in\mathbb{R}^+$ controls the strength of the smoothing, as illustrated in Fig.~\ref{fig:visualization_d}.

\subsubsection{Image Classification and Wrap-up}
The last step detects graph properties by an image classifier $f_{\boldsymbol{\theta}_c}(\boldsymbol{\mathsf{M}}) \in \{0,1\}$, which can be, for example, ResNet-50 and ViT, with a set of learnable parameters denoted as $\boldsymbol{\theta}_c$. The proposed pipeline can be summarized as
\begin{align}
\VSAL_{\thetaGen, \boldsymbol{\theta}_c}(G,\mathbf{z}) \in \{0,1\},
\end{align}
where it takes the graph $G$ and the latent noise vector $\mathbf{z}$ as input with learnable parameters being $\big\{\thetaGen, \boldsymbol{\theta}_c\big\}$.

\subsection{Training Method}
\label{subsec:training}
This section describes the training procedure of VSAL, i.e., deciding the best set of parameters $\big\{\thetaGen, \boldsymbol{\theta}_c\big\}$. The parameter $ \boldsymbol{\theta}_c$ associated with the image classifier can be straightforwardly optimized by minimizing the classification loss, but finding the best $\thetaGen$ is less trivial. This is because the layouts from a random set $\thetaGen$ do not exhibit enough structural information for graph property detection. To address such issues, instead of training the generator from scratch, we seek to guide the generator to produce layouts that are similar to the given principled layouts (i.e., reference layouts), which is achieved through adversarial training; meanwhile, the parameters $\thetaGen$ are also optimized towards classification accuracy. In the rest of this section, we will first present the selected reference layouts and then describe the training method. 

\subsubsection{Reference Layout}
In generating the reference layout of a given graph $G=(V,E,\mathbf{A})$, an initial layout $\mathbf{L}_\text{initial}\in\mathbb{R}^{|V|\times 2}$ is first created based on principled designs. We explore three principled designs and use the uniform layout as a baseline. For a graph $G$, the position of each node $i\in V$ is determined as follows.
\begin{itemize}
    \item Circular \cite{bhavsar2022graph}: The nodes are arranged uniformly on a circle of radius 
$r_1\in\mathbb{R^+}$: 
\begin{align*}
    \mathbf{L}_{\text{initial}}[i, :] =  \big( r_1  \cos(2\pi i/|V| ), r _1 \sin (2\pi i/|V| ) \big).
\end{align*}
    \item Spiral \cite{carlis1998interactive}: The nodes are placed along an Archimedes spiral path parameterized by the offset factor $r_2 \in\mathbb{R^+}$:
\begin{align*} 
\mathbf{L}_{\text{initial}}[i, :] = \big(i \cos(i \cdot r_2),\, i \sin(i \cdot r_2)\big).
\end{align*} 
    \item Shell \cite{diaz2002survey}: 
    The nodes are grouped into $S \in\mathbb{Z}^+$ concentric shells (circles), where shell $j$ has radius $r_j \in \mathbb{R}^+$ and contains $n_j$ nodes such that $\sum_{j=1}^S n_j = |V|$: 
\begin{align*}
    \mathbf{L}_{\text{initial}}[i, :] = \big(r_j \cos(2\pi i / n_j), r_j \sin(2\pi i / n_j)\big).
\end{align*}
\item Uniform: The nodes are positioned randomly within a layout region bounded by $b\in\mathbb{R}^+$: 
\begin{align*}
    \mathbf{L}_{\text{initial}}[i, :] = (x_i, y_i), \quad x_i, y_i \sim U(-b, b).
\end{align*}
\end{itemize}

\begin{figure}[t!]
\centering
\subfloat{\includegraphics[width=0.10\textwidth]{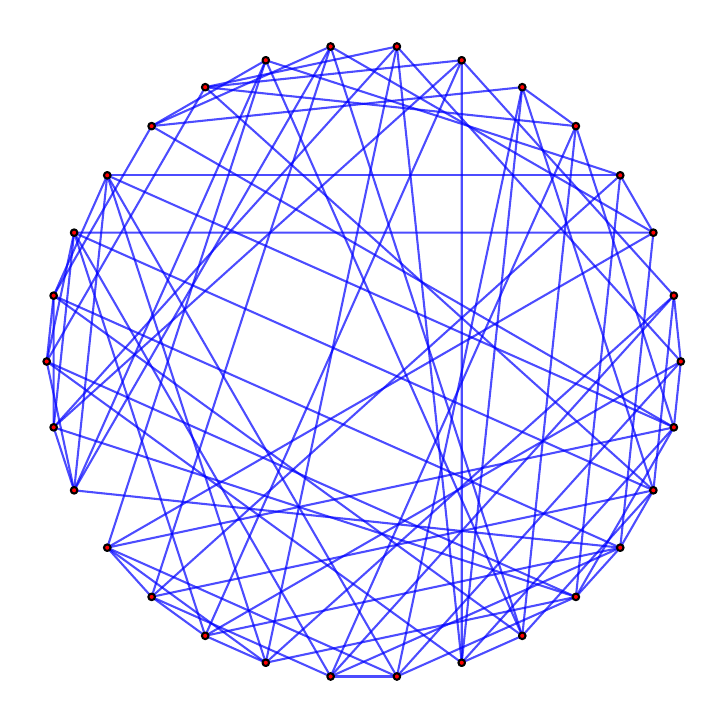}}\hspace{2mm}
\subfloat{\includegraphics[width=0.10\textwidth]{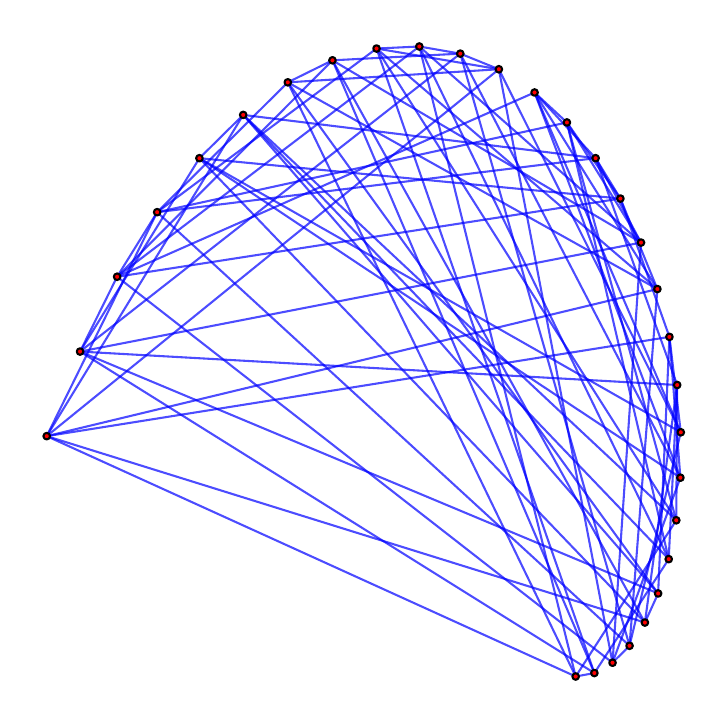}}\hspace{2mm}
\subfloat{\includegraphics[width=0.10\textwidth]{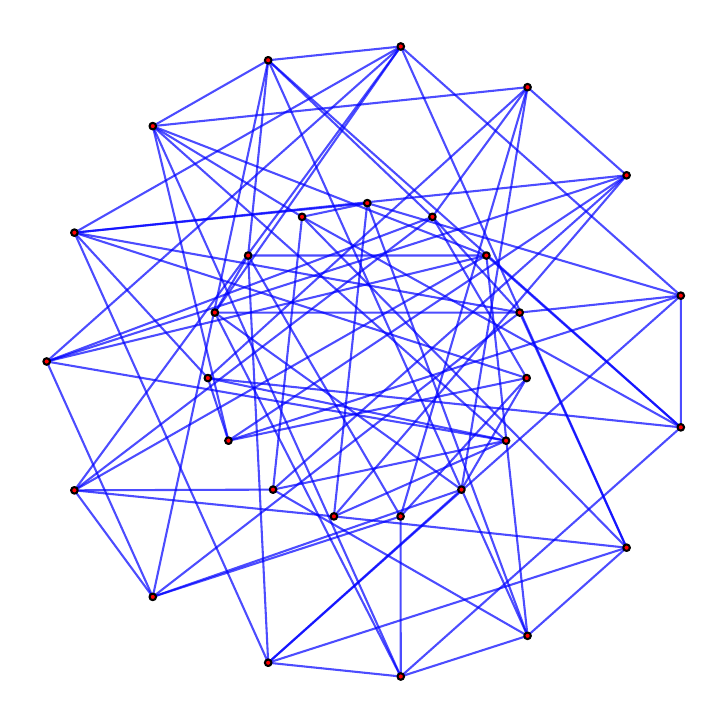}}\hspace{2mm}
\subfloat{\includegraphics[width=0.10\textwidth]{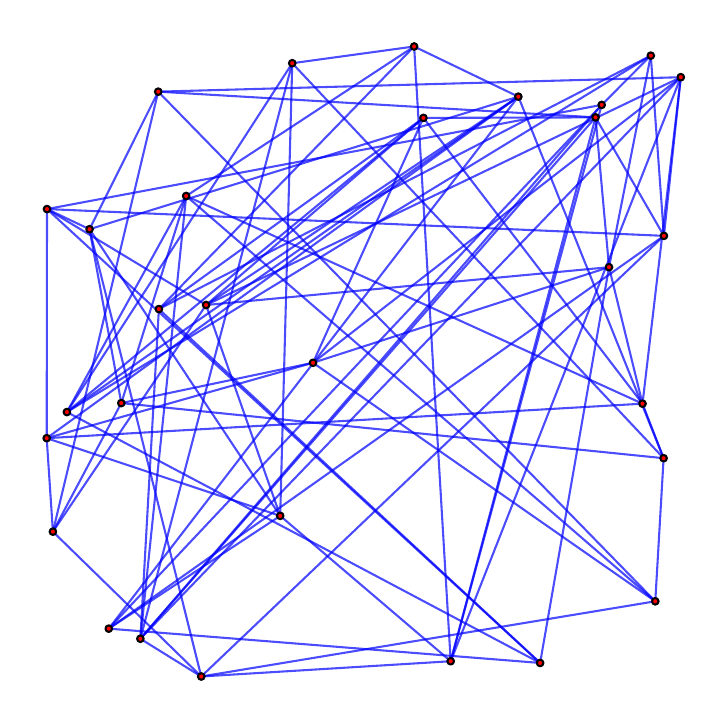}}
\setcounter{subfigure}{0} 
\renewcommand{\thesubfigure}{\alph{subfigure}} 
\subfloat[Circular]{\label{fig:1-2}\includegraphics[width=0.10\textwidth]{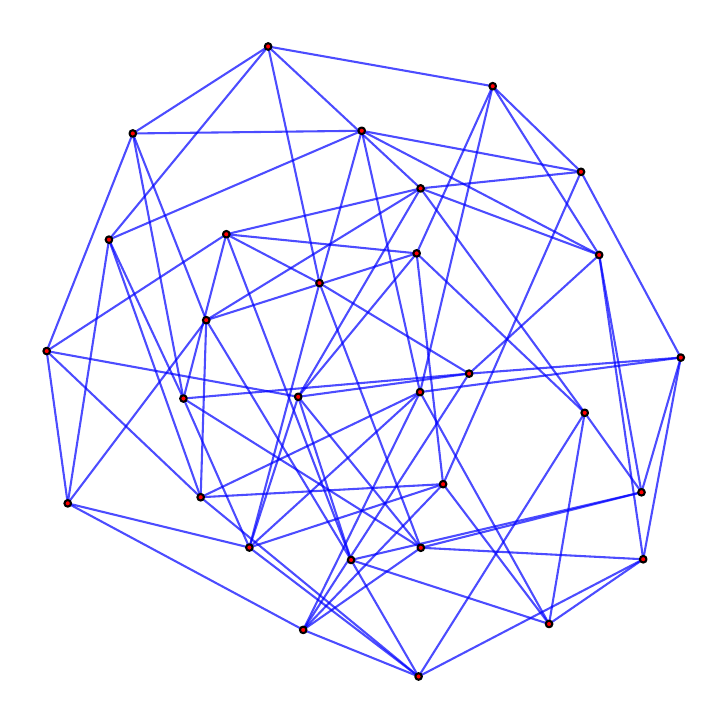}}\hspace{2mm}
\subfloat[Spiral]{\label{fig: 1-4}\includegraphics[width=0.10\textwidth]{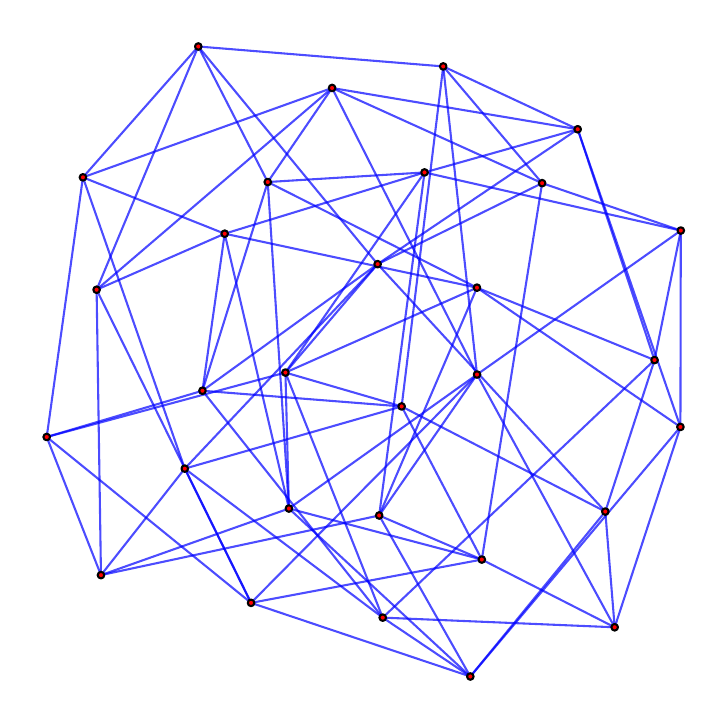}}\hspace{2mm}
\subfloat[Shell]{\label{fig: 1-6}\includegraphics[width=0.10\textwidth]{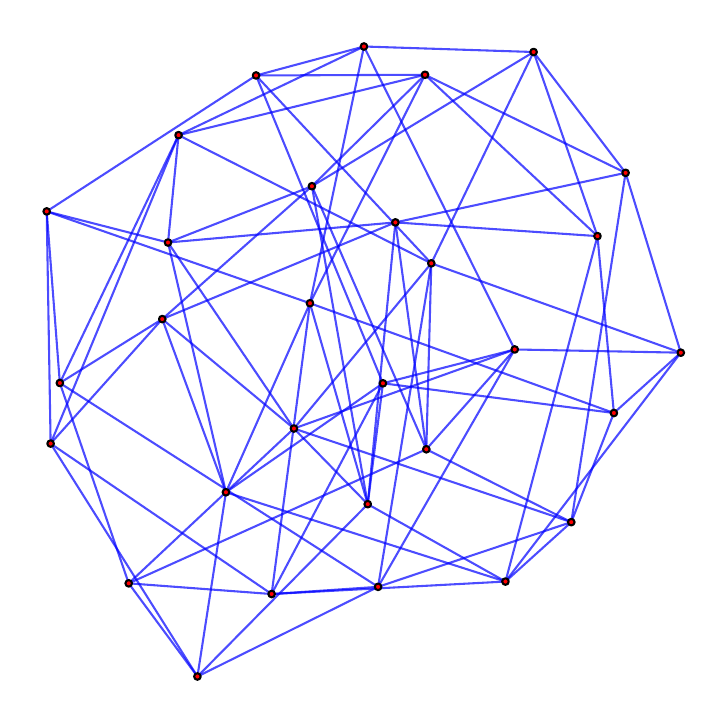}}\hspace{2mm}
\subfloat[Uniform]{\label{fig: 1-8}\includegraphics[width=0.10\textwidth]{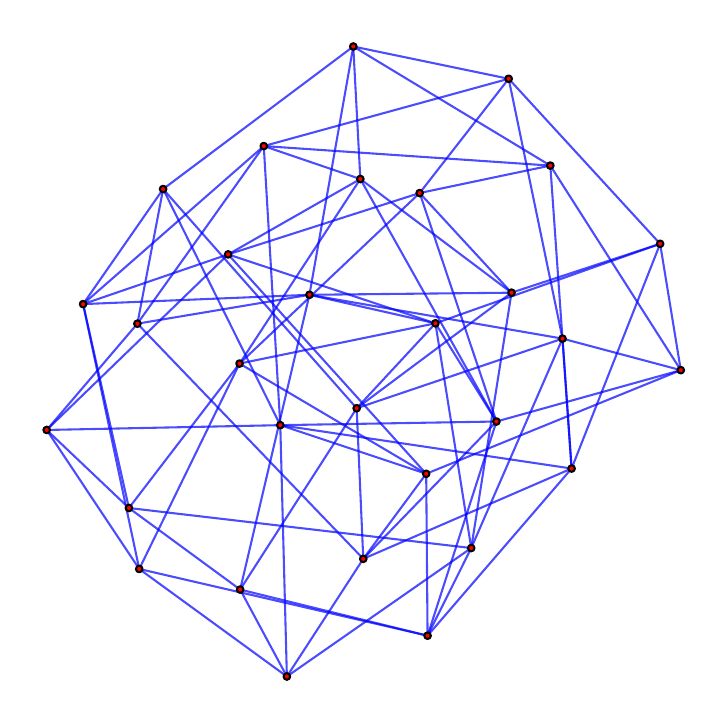}}
\Description{This figure shows the role of refinement, which can reduce the edge overlappings of the graph layouts.}
\caption{Each pair shows an example of the initial layout (top) and the refined layout (bottom) via Eq.~\ref{eq: refine}.}
\label{fig:visualization_layouts}
\end{figure}

To further refine the layout and reduce edge crossings, we apply a combination of the spring algorithm \cite{fruchterman1991graph} and the Kamada-Kawai algorithm \cite{kamada1989algorithm},
of which the details can be found in Appendix~\ref{apd:layout_algorithm}. 
The resulting reference layout is denoted as
\begin{align}
\label{eq: refine}
    \rref(G) = \Kamada\big(\Spring(\mathbf{L}_{\text{initial}})\big) \in \mathbb{R}^{|V|\times 2}.
\end{align}
Fig.~\ref{fig:visualization_layouts} intuitively shows that this refinement can effectively reduce edge crossings and potentially improve the quality of initial layouts, 
thereby facilitating the training of the generator, as evidenced in our ablation studies (Sec.~\ref{subsec:Ablation}).

\subsubsection{Adversarial Training with Classification Feedback}
To encourage the generator to produce layouts that resemble reference layouts, we employ a Wasserstein adversarial training strategy \cite{arjovsky2017wasserstein}.
To this end, 
we design a class $\mathcal{D}_\text{dis}$ of discriminators to distinguish between generated layouts $g_{\boldsymbol{\theta_\text{gen}}}(G,\mathbf{z})$ (Eq.~\ref{eq: generator}) and reference layouts $\rref(G)$ (Eq.~\ref{eq: refine}), where each discriminator $d_{\boldsymbol{\theta}_\text{dis}}$ is parameterized by $\boldsymbol{\theta}_\text{dis}=\big\{\theta_\text{dis}^{1},\theta_\text{dis}^{2}\big\}$:
\begin{align*}
    \mathcal{D}_\text{dis} \subseteq \Big\{d_{\boldsymbol{\theta}_\text{dis}}(\mathbf{L},G=(V,E,\mathbf{A})):\mathbb{R}^{|V|\times 2} \times \mathcal{G} \rightarrow \mathbb{R} \Big\}.
\end{align*}
In particular, the input graph layout $\mathbf{L}$ (either $g_{\boldsymbol{\theta_\text{gen}}}(G,\mathbf{z})$ or $\rref(G)$) and the adjacency matrix $\mathbf{A}$ are processed via a graph encoder, similar to the one used in the generator, to extract structural features.
\begin{align*}
    H_D &= \text{GraphEncoder}_{\theta_\text{dis}^{1}}(\mathbf{L}, \mathbf{A}) \in \mathbb{R}^{|V| \times d_s},
\end{align*}
where $d_s\in \mathbb{Z}^+$ is the hidden dimension and $\theta_\text{dis}^{1}$ is a set of learnable parameters. After that, a global average pooling layer is used to aggregate node features $H_D$ into a graph-level representation.
\begin{align*}    
\mathbf{h}_{\text{graph}} = \mean\big(H_D\big)  \in \mathbb{R}^{d_s}.
\end{align*} 
Finally, $\mathbf{h}_{\text{graph}}$ is passed through an MLP parameterized by $\theta_\text{dis}^{2}$ to output the validity score, where higher scores indicate that the input layout $\mathbf{L}$ is more similar to the reference layout.
\begin{align*}
     d_{\boldsymbol{\theta}_\text{dis}}(\mathbf{L},G)= \text{MLP}_{\theta_\text{dis}^{2}}(\mathbf{h}_{\text{graph}}) \in \mathbb{R}.
\end{align*}

Now, we are ready to discuss the training process. Given the training set $\big\{(G_i, y_i)\big\}_{i=1}^n$, a collection $\big\{\mathbf{z}_i^j\big\}_{j=1}^{m}$ of $m \in \mathbb{Z}^+$ noise vectors are sampled from $P_{\mathcal{Z}}$ for each pair $(G_i,y_i)$. 

\textbf{Training of $\boldsymbol{\theta}_\text{gen}$.} The generator $g_{\boldsymbol{\theta}_\text{gen}}$ is trained to produce graph layouts that can receive high scores from the discriminator $d_{\boldsymbol{\theta}_\text{dis}}$ while also improving the detection effect. This is achieved by minimizing a combination of the empirical adversarial loss and classification loss:
\begin{align}
\nonumber
&\hat{\mathcal{L}}_{g}\big(\boldsymbol{\theta}_\text{gen},\boldsymbol{\theta}_\text{dis},\boldsymbol{\theta}_c\big) \define \hat{\mathcal{L}}_{\text{adv}}\big( \boldsymbol{\theta}_\text{gen}, \boldsymbol{\theta}_\text{dis} \big) + \lambda_{c} \hat{\mathcal{L}}_{c}\big(  \boldsymbol{\theta}_\text{gen},\boldsymbol{\theta}_c\big),\\
\nonumber
 &\hat{\mathcal{L}}_{\text{adv}}\big(\boldsymbol{\theta}_\text{gen}, \boldsymbol{\theta}_\text{dis}  \big) \define -  \frac{1}{nm} \sum_{i=1}^{n} \sum_{j=1}^m d_{\boldsymbol{\theta}_\text{dis}}\Big(g_{\boldsymbol{\theta}_\text{gen}}\big( G_{i},\mathbf{z}_{i}^j\big),G_{i}\Big),\\
 \nonumber
&\hat{\mathcal{L}}_c\big(\boldsymbol{\theta}_\text{gen} ,\boldsymbol{\theta}_c\big) \define \frac{1}{nm} \sum_{i=1}^{n}\sum_{j=1}^m  \ell_c\left( \VSAL_{\thetaGen, \boldsymbol{\theta}_c}\big(G_{i},\mathbf{z}_{i}^j\big), y_{i} \right),
\end{align}
where $\ell_c$ is the cross-entropy loss function and $\lambda_{c}\in \mathbb{R}^+$ is a weight factor that balances these two loss components.

\textbf{Training of $\boldsymbol{\theta}_\text{dis}$.} The discriminator is trained to differentiate between generated layouts and reference layouts by optimizing the Wasserstein loss with gradient penalty \cite{gulrajani2017improved}:
\begin{align}
\nonumber
    \hat{\mathcal{L}}_{d}\big(\boldsymbol{\theta}_\text{gen}, \boldsymbol{\theta}_\text{dis}\big) &\define -\frac{1}{nm}\sum_{i=1}^n \sum_{j=1}^m d_{\boldsymbol{\theta}_\text{dis}}\Big(g_{\boldsymbol{\theta}_\text{gen}}\big(G_{i}, \mathbf{z}_{i}^j\big),G_{i} \Big) \\ 
   + \frac{1}{n}\sum_{i=1}^n &d_{\boldsymbol{\theta}_\text{dis}}\big(\rref(G_i),G_{i} \big)  +\lambda_{\text{gp}}\hat{\mathcal{L}}_{\text{gp}}\big(\boldsymbol{\theta}_\text{gen}, \boldsymbol{\theta}_\text{dis} \big), 
   \nonumber
\end{align}
where 
$\lambda_{\text{gp}}\in\mathbb{R}^+$ is the gradient penalty coefficient. The gradient penalty term $\hat{\mathcal{L}}_{\text{gp}}$ enforces the 1-Lipschitz constraint to prevent vanishing or exploding gradients in the discriminator, ensuring that the generator receives consistently effective guidance \cite{arjovsky2017wasserstein,gulrajani2017improved}:
\begin{align*}
    &\hat{\mathcal{L}}_{\text{gp}}\big(\boldsymbol{\theta}_\text{gen}, \boldsymbol{\theta}_\text{dis}\big) \define \frac{1}{nm} \sum_{i=1}^{n} \sum_{j=1}^m \left( \big\|\nabla_{\hat{\mathbf{L}}_{i}^j} d_{\boldsymbol{\theta}_\text{dis}}\big(\hat{\mathbf{L}}_{i}^j,G_{i} \big)\big\|_2 - 1 \right)^2,\\
    &\hat{\mathbf{L}}_{i}^j = \lambda \rref(G_i) + (1-\lambda)g_{\boldsymbol{\theta}_\text{gen}}\big(G_{i}, \mathbf{z}_{i}^j\big), \, \lambda \sim U(0,1),
\end{align*}
where $\hat{\mathbf{L}}_{i}^j $ is the interpolation between the reference layout $\rref(G_i)$ and the generated layout $g_{\boldsymbol{\theta}_\text{gen}}\big(G_{i}, \mathbf{z}_{i}^j\big)$.

\textbf{Training of $\boldsymbol{\theta}_\text{c}$.} The classifier is trained by minimizing the classification loss $\hat{\mathcal{L}}_c\big(\boldsymbol{\theta}_\text{gen} ,\boldsymbol{\theta}_c\big)$.

\textbf{Optimization scheme.} 
To stabilize VSAL training for better detection accuracy, we pretrain $\boldsymbol{\theta}_\text{gen}$ and $\boldsymbol{\theta}_\text{dis}$ to mitigate the instability of adversarial training \cite{thanhimproving}, which is done by
\begin{align}
\label{eq: pre_train}
    \pdv{\hat{\mathcal{L}}_{\text{adv}}\big( \boldsymbol{\theta}_\text{gen} , \boldsymbol{\theta}_\text{dis} \big)}{\boldsymbol{\theta}_\text{gen}}~ \text{and}~   \pdv{\hat{\mathcal{L}}_{d}\big(\boldsymbol{\theta}_\text{gen}, \boldsymbol{\theta}_\text{dis}\big) }{\boldsymbol{\theta}_\text{dis}}.
\end{align}
This pretraining enables the generator to learn the fundamental principles for generating structured layouts,
while ensuring that the discriminator can effectively differentiate reference layouts from poorly structured ones, thus providing a stable foundation for the training of VSAL.
After pretraining, we iteratively optimize the parameters of each component through coordinated gradient descent via
\begin{align*}
\pdv{\hat{\mathcal{L}}_{g}\big(\boldsymbol{\theta}_\text{gen},\boldsymbol{\theta}_\text{dis},\boldsymbol{\theta}_c\big)}{\boldsymbol{\theta}_\text{gen}}, \pdv{\hat{\mathcal{L}}_{d}\big(\boldsymbol{\theta}_\text{gen}, \boldsymbol{\theta}_\text{dis} \big) }{\boldsymbol{\theta}_\text{dis}}, ~ \text{and},~  \pdv{\hat{\mathcal{L}}_c\big(\boldsymbol{\theta}_\text{gen} ,\boldsymbol{\theta}_c\big)}{\boldsymbol{\theta}_c}.    
\end{align*}

\begin{table}[!pt] 
\centering \caption{Statistics of datasets. Graph size: 
small (4-20 nodes), medium (21-50 nodes), large (401-500 nodes), and huge (901-1000 nodes). ``T'' and ``F'' denote true and false, respectively.} 
\label{tab:dataset} 
\begin{tabular}{@{}ccccccccc@{}}
        \toprule
         & \multicolumn{4}{c}{House of Graphs} & \multicolumn{4}{c}{Synthetic}\\
         \cmidrule(lr){2-5}  \cmidrule(lr){6-9}
        & \multicolumn{2}{c}{Small} & \multicolumn{2}{c}{Medium} &   \multicolumn{2}{c}{Large} &  \multicolumn{2}{c}{Huge}\\
        \cmidrule(lr){2-3} \cmidrule(lr){4-5} \cmidrule(lr){6-7} \cmidrule(lr){8-9}
        & T & F & T & F & T & F & T & F\\
        \midrule
        Ham &2277  &1838  &7453  &5739   &2000  &2000 &2000  &2000 \\
        Planar &1947  &2000  &1999  &2000   &2000  &2000 &2000  &2000 \\
        Claw &919  &946  &-  &-   &2000  &2000 &2000  &2000 \\
        Tree &213  &481  &-  &-   &2000  &2000 &2000  &2000 \\
        \bottomrule
    \end{tabular}
\end{table}

\section{Experiments}
\label{sec:experiment}
In this section, we evaluate the effectiveness of our VSAL framework and demonstrate that VSAL behaves in the designed way. 

\subsection{Settings}
\subsubsection{Tasks and Datasets}
We consider four well-known graph property detection tasks: Hamiltonian cycle problem (\textbf{Ham}) \cite{gould2014recent}, planarity verification (\textbf{Planar}) \cite{hopcroft1974efficient}, claw-free graph classification (\textbf{Claw}) \cite{faudree1997claw}, and tree recognition (\textbf{Tree}) \cite{shasha2002algorithmics}; their formal definitions can be found in Appendix~\ref{apd:tasks_definition}. 
We conduct experiments on datasets of varying graph sizes. The small and medium datasets are borrowed from the House of Graphs \cite{coolsaet2023house}, while the large and huge datasets are synthetically produced, as described in Appendix~\ref{apd:dataset}. The summary statistics for all datasets are provided in Table~\ref{tab:dataset}.


\subsubsection{VSAL Settings}
To comprehensively evaluate VSAL, we explore multiple configurations of graph encoders, backbone classifiers, and input resolutions. Specifically, we implement VSAL with two graph encoders: DenseGCN and Graphormer. As for image classification, we adopt three widely used backbone classifiers: ResNet-50, Vision Transformer (ViT), and EfficientNet \cite{tan2019efficientnet}, all initialized with ImageNet-pretrained weights \cite{he2016deep,dosovitskiy2021an}.  
For ResNet-50 and ViT, we use an input resolution of $224 \times 224$. For EfficientNet, we evaluate three resolutions: $224 \times 224$, $380 \times 380$, and $528 \times 528$. We denote these model variants as \textbf{VSAL-R-224}, \textbf{VSAL-V-224}, \textbf{VSAL-E-224}, \textbf{VSAL-E-380}, and \textbf{VSAL-E-528}, respectively.

\subsubsection{Baseline Methods}
We compare VSAL with the following baseline methods, which have been introduced in Sec.~\ref{sec:related}.
\begin{itemize}
    \item \textbf{VN-Solver} \cite{samizadeh2023vn}: We follow the official implementation and use both circular and spiral layouts with a ResNet-50 backbone at $224 \times 224$ resolution. 
    \item \textbf{Graphormer} \cite{ying2021transformers} We use the offical pretrained checkpoint pcqm4mv2\_graphormer\_base,  configured with $768$ hidden units and $12$ attention heads, and conduct fine-tuning.
    \item \textbf{Graphormer-GD} \cite{zhangrethinking}: We train the model from scratch following the official codebase.
    \item \textbf{EquiformerV2} \cite{liaoequiformerv2}:We load the publicly released checkpoint eq2\_83M\_2M.pt and fine-tune it on our datasets. 
    \item \textbf{GraphsGPT} \cite{gao2024graph}: We adopt the official released encoder checkpoint GraphsGPT-4W and fine-tune it for our tasks. 
    \item \textbf{Random}: A baseline assigning labels uniformly at random.
\end{itemize}

\begin{table*}[t!]
\small
\renewcommand{\arraystretch}{1.1} 
    \centering
    \caption{F1 scores of all models on four graph property detection tasks. Each model is trained with a size of 200 samples on small tree graphs and 1000 samples on all other datasets. VSAL is implemented with two representative graph encoders: DensedGCN and Graphormer. All VSAL variants use circular initial layouts.  VN-Solver-C and VN-Solver-S denote VN-Solver with circular and spiral layouts, respectively. The top three results for each task are highlighted.
    }
    \label{tab:F1 Hamiltonian Cycle Problem_large_huge}
    \begin{tabular}{@{}l@{\hskip 2pt}c@{\hskip 6pt}c@{\hskip 12pt}c@{\hskip 2.5pt}c@{\hskip 2.5pt}c@{\hskip 2.5pt}c@{\hskip 6pt}c@{\hskip 2.5pt}c@{\hskip 2.5pt}c@{\hskip 2.5pt}c@{\hskip 6pt}c@{\hskip 2.5pt}c@{\hskip 2.5pt}c@{\hskip 6pt}c@{\hskip 2.5pt}c@{\hskip 2.5pt}c@{}}
        \toprule
       && &\multicolumn{4}{c}{Ham} & \multicolumn{4}{c}{Planar}  & \multicolumn{3}{c}{Claw}  & \multicolumn{3}{c}{Tree}\\
        \cmidrule(lr){4-7} \cmidrule(lr){8-11} \cmidrule(lr){12-14}\cmidrule(lr){15-17}  
        ~ & & &Small &Medium & Large &Huge & Small &Medium & Large &Huge & Small & Large &Huge & Small & Large &Huge   \\
        \midrule
        \multirow{10}{*}{VSAL} &\multirow{5}{*}{DenseGCN} & R-224 & $0.89 \text{\tiny (0.01)}$  & $\boldsymbol{0.96} \text{\tiny (0.01)}$ & $0.89 \text{\tiny (0.00)}$ & $0.89 \text{\tiny (0.01)}$ & $0.86 \text{\tiny (0.01)}$ & $\boldsymbol{0.97} \text{\tiny (0.01)}$ & $0.91 \text{\tiny (0.01)}$ & $0.89 \text{\tiny (0.01)}$ & $0.91 \text{\tiny (0.01)}$ & $0.91 \text{\tiny (0.01)}$ & $0.90 \text{\tiny (0.01)}$ & $0.93 \text{\tiny (0.01)}$ & $0.91 \text{\tiny (0.01)}$ & $0.90 \text{\tiny (0.01)}$  \\
         & &V-224  & $0.90 \text{\tiny (0.01)}$  & $\boldsymbol{0.98} \text{\tiny (0.00)}$ & $0.91 \text{\tiny (0.00)}$ & $0.90 \text{\tiny (0.01)}$ & $0.89 \text{\tiny (0.01)}$ & $\boldsymbol{0.97} \text{\tiny (0.00)}$ & $0.92 \text{\tiny (0.01)}$ & $0.91 \text{\tiny (0.01)}$ & $0.93 \text{\tiny (0.01)}$ & $\boldsymbol{0.93} \text{\tiny (0.01)}$ & $\boldsymbol{0.92} \text{\tiny (0.01)}$ & $\boldsymbol{0.96} \text{\tiny (0.01)}$ & $0.94 \text{\tiny (0.01)}$ & $0.92 \text{\tiny (0.00)}$  \\
         & &E-224 & $0.89 \text{\tiny (0.01)}$  & $\boldsymbol{0.97} \text{\tiny (0.00)}$ & $0.89 \text{\tiny (0.01)}$ & $0.88 \text{\tiny (0.01)}$ & $0.87 \text{\tiny (0.01)}$ & $\boldsymbol{0.97} \text{\tiny (0.00)}$ & $0.91 \text{\tiny (0.01)}$ & $0.90 \text{\tiny (0.01)}$ & $0.92 \text{\tiny (0.01)}$ & $0.91 \text{\tiny (0.01)}$ & $0.91 \text{\tiny (0.01)}$ & $\boldsymbol{0.96} \text{\tiny (0.01)}$ & $0.91 \text{\tiny (0.01)}$ & $0.90 \text{\tiny (0.01)}$  \\
          & & E-380& $0.90 \text{\tiny(0.01)}$  & $\boldsymbol{0.97} \text{\tiny (0.01)}$ & $0.92 \text{\tiny (0.00)}$ & $\boldsymbol{0.91} \text{\tiny (0.00)}$ & $0.87 \text{\tiny (0.01)}$ & $\boldsymbol{0.97} \text{\tiny (0.00)}$ & $\boldsymbol{0.93} \text{\tiny (0.01)}$ & $0.91 \text{\tiny (0.01)}$ & $0.92 \text{\tiny (0.01)}$ & $\boldsymbol{0.92} \text{\tiny (0.01)}$ & $0.91 \text{\tiny (0.00)}$ & $\boldsymbol{0.96} \text{\tiny (0.01)}$ & $0.93 \text{\tiny (0.01)}$ & $\boldsymbol{0.93} \text{\tiny (0.01)}$  \\
        & & E-528 & $\boldsymbol{0.91} \text{\tiny (0.01)}$  & $\boldsymbol{0.98} \text{\tiny (0.01)}$ & $\boldsymbol{0.94} \text{\tiny (0.01)}$ & $\boldsymbol{0.92} \text{\tiny (0.00)}$ & $0.87 \text{\tiny (0.01)}$ & $\boldsymbol{0.98} \text{\tiny (0.01)}$ & $\boldsymbol{0.93} \text{\tiny (0.00)}$ & $\boldsymbol{0.92} \text{\tiny (0.01)}$ & $0.93 \text{\tiny (0.01)}$ & $\boldsymbol{0.93} \text{\tiny (0.01)}$ & $\boldsymbol{0.92} \text{\tiny (0.01)}$ & $\boldsymbol{0.96} \text{\tiny (0.01)}$ & $\boldsymbol{0.95} \text{\tiny (0.01)}$ & $\boldsymbol{0.94} \text{\tiny (0.01)}$  \\
        \cmidrule(lr){2-17}
        &\multirow{5}{*}{Graphormer} & R-224& $\boldsymbol{0.93} \text{\tiny (0.01)}$& $\boldsymbol{0.97} \text{\tiny (0.00)}$& $\boldsymbol{0.93} \text{\tiny (0.01)}$& $0.90 \text{\tiny (0.01)}$& $\boldsymbol{0.90} \text{\tiny (0.01)}$& $\boldsymbol{0.97} \text{\tiny (0.01)}$& $0.92 \text{\tiny (0.00)}$& $0.91 \text{\tiny (0.01)}$& $\boldsymbol{0.94} \text{\tiny (0.01)}$& $\boldsymbol{0.93} \text{\tiny (0.01)}$& $0.91 \text{\tiny (0.01)}$& $\boldsymbol{0.97} \text{\tiny (0.01)}$& $\boldsymbol{0.95} \text{\tiny (0.01)}$& $0.92 \text{\tiny (0.01)}$    \\
         & & V-224 & $\boldsymbol{0.93} \text{\tiny (0.01)}$& $\boldsymbol{0.98} \text{\tiny (0.00)}$& $\boldsymbol{0.94} \text{\tiny (0.01)}$& $\boldsymbol{0.92} \text{\tiny (0.01)}$& $\boldsymbol{0.92} \text{\tiny (0.01)}$& $\boldsymbol{0.98} \text{\tiny (0.00)}$& $\boldsymbol{0.94} \text{\tiny (0.01)}$& $\boldsymbol{0.93} \text{\tiny (0.01)}$& $\boldsymbol{0.95} \text{\tiny (0.01)}$& $\boldsymbol{0.95} \text{\tiny (0.01)}$& $\boldsymbol{0.94} \text{\tiny (0.01)}$& $\boldsymbol{0.98} \text{\tiny (0.01)}$& $\boldsymbol{0.96} \text{\tiny (0.01)}$& $\boldsymbol{0.94} \text{\tiny (0.01)}$ \\
         & & E-224  & $\boldsymbol{0.93} \text{\tiny (0.01)}$& $\boldsymbol{0.98} \text{\tiny (0.01)}$& $\boldsymbol{0.93} \text{\tiny (0.01)}$& $\boldsymbol{0.91} \text{\tiny (0.01)}$& $\boldsymbol{0.91} \text{\tiny (0.01)}$& $\boldsymbol{0.98} \text{\tiny (0.00)}$& $\boldsymbol{0.93} \text{\tiny (0.01)}$& $\boldsymbol{0.92} \text{\tiny (0.01)}$& $\boldsymbol{0.94} \text{\tiny (0.00)}$& $\boldsymbol{0.93} \text{\tiny (0.01)}$& $\boldsymbol{0.92} \text{\tiny (0.01)}$& $\boldsymbol{0.98} \text{\tiny (0.01)}$& $\boldsymbol{0.95} \text{\tiny (0.00)}$& $\boldsymbol{0.93} \text{\tiny (0.01)}$ \\
         & & E-380 & $\boldsymbol{0.93} \text{\tiny (0.01)}$& $\boldsymbol{0.98} \text{\tiny (0.00)}$& $\boldsymbol{0.94} \text{\tiny (0.00)}$& $\boldsymbol{0.93} \text{\tiny (0.01)}$& $\boldsymbol{0.91} \text{\tiny (0.01)}$& $\boldsymbol{0.98} \text{\tiny (0.00)}$& $\boldsymbol{0.94} \text{\tiny (0.01)}$& $\boldsymbol{0.92} \text{\tiny (0.01)}$& $\boldsymbol{0.95} \text{\tiny (0.01)}$& $\boldsymbol{0.93} \text{\tiny (0.01)}$& $\boldsymbol{0.93} \text{\tiny (0.01)}$& $\boldsymbol{0.98} \text{\tiny (0.00)}$& $\boldsymbol{0.96} \text{\tiny (0.01)}$& $\boldsymbol{0.94} \text{\tiny (0.01)}$ \\
         & & E-528 & $\boldsymbol{0.94} \text{\tiny (0.00)}$& $\boldsymbol{0.98} \text{\tiny (0.01)}$& $\boldsymbol{0.95} \text{\tiny (0.01)}$& $\boldsymbol{0.93} \text{\tiny (0.00)}$& $\boldsymbol{0.92} \text{\tiny (0.01)}$& $\boldsymbol{0.99} \text{\tiny (0.01)}$& $\boldsymbol{0.95} \text{\tiny (0.00)}$& $\boldsymbol{0.94} \text{\tiny (0.01)}$& $\boldsymbol{0.96} \text{\tiny (0.01)}$& $\boldsymbol{0.95} \text{\tiny (0.00)}$& $\boldsymbol{0.94} \text{\tiny (0.01)}$& $\boldsymbol{0.98} \text{\tiny (0.00)}$& $\boldsymbol{0.97} \text{\tiny (0.01)}$& $\boldsymbol{0.95} \text{\tiny (0.01)}$ \\
        \midrule
         \multicolumn{3}{c}{VN-Solver-C \cite{samizadeh2023vn}} & $0.83 \text{\tiny (0.03)}$  & $0.95 \text{\tiny (0.01)}$ & $0.57 \text{\tiny (0.03)}$ & $0.53 \text{\tiny (0.01)}$ & $0.85 \text{\tiny (0.02)}$ & $0.95 \text{\tiny (0.01)}$ & $0.51 \text{\tiny (0.02)}$ & $0.51 \text{\tiny (0.01)}$ & $0.86 \text{\tiny (0.01)}$ & $0.60 \text{\tiny (0.01)}$ & $0.58 \text{\tiny (0.02)}$ & $0.83 \text{\tiny (0.02)}$ & $0.76 \text{\tiny (0.01)}$ & $0.64 \text{\tiny (0.02)}$  \\
         \multicolumn{3}{c}{VN-Solver-S \cite{samizadeh2023vn}} & $0.80 \text{\tiny (0.02)}$  & $0.95 \text{\tiny (0.01)}$ & $0.81 \text{\tiny (0.01)}$ & $0.76 \text{\tiny (0.01)}$ & $0.84 \text{\tiny (0.01)}$ & $0.94 \text{\tiny (0.02)}$ & $0.53 \text{\tiny (0.01)}$ & $0.51 \text{\tiny (0.02)}$ & $0.85 \text{\tiny (0.01)}$ & $0.68 \text{\tiny (0.02)}$ & $0.62 \text{\tiny (0.02)}$ & $0.81 \text{\tiny (0.03)}$ & $0.77 \text{\tiny (0.02)}$ & $0.68 \text{\tiny (0.02)}$  \\
         \multicolumn{3}{c}{Graphormer \cite{ying2021transformers}} & $0.88 \text{\tiny (0.00)}$  & $0.93 \text{\tiny (0.00)}$ & $0.88 \text{\tiny (0.00)}$ & $0.80 \text{\tiny (0.00)}$ & $0.88 \text{\tiny (0.00)}$ & $0.96 \text{\tiny (0.00)}$ & $0.90 \text{\tiny (0.00)}$ & $0.83 \text{\tiny (0.00)}$ & $0.89 \text{\tiny (0.00)}$ & $0.80 \text{\tiny (0.00)}$ & $0.69 \text{\tiny (0.00)}$ & $\boldsymbol{0.98} \text{\tiny (0.00)}$ & $0.87 \text{\tiny (0.00)}$ & $0.79 \text{\tiny (0.00)}$  \\
         \multicolumn{3}{c}{Graphormer-GD \cite{zhangrethinking}} & $0.88 \text{\tiny (0.01)}$  & $0.95 \text{\tiny (0.01)}$ & $0.90 \text{\tiny (0.00)}$ & $0.82 \text{\tiny (0.01)}$ & $0.89 \text{\tiny (0.01)}$ & $0.95 \text{\tiny (0.01)}$ & $0.89 \text{\tiny (0.01)}$ & $0.82 \text{\tiny (0.01)}$ & $0.90 \text{\tiny (0.02)}$ & $0.83 \text{\tiny (0.02)}$ & $0.71 \text{\tiny (0.01)}$ & $\boldsymbol{0.98} \text{\tiny (0.00)}$ & $0.88 \text{\tiny (0.01)}$ & $0.81 \text{\tiny (0.01)}$  \\
         \multicolumn{3}{c}{EquiformerV2 \cite{liaoequiformerv2}} & $0.85 \text{\tiny (0.01)}$  & $0.92 \text{\tiny (0.01)}$ & $0.89 \text{\tiny (0.01)}$ & $0.79 \text{\tiny (0.01)}$ & $0.83 \text{\tiny (0.00)}$ & $0.94 \text{\tiny (0.00)}$ & $0.85 \text{\tiny (0.01)}$ & $0.79 \text{\tiny (0.02)}$ & $0.85 \text{\tiny (0.01)}$ & $0.77 \text{\tiny (0.02)}$ & $0.67 \text{\tiny (0.02)}$ & $0.94 \text{\tiny (0.00)}$ & $0.85 \text{\tiny (0.02)}$ & $0.77 \text{\tiny (0.01)}$  \\
         \multicolumn{3}{c}{GraphsGPT \cite{gao2024graph}} & $0.87 \text{\tiny (0.00)}$  & $0.94 \text{\tiny (0.01)}$ & $0.90 \text{\tiny (0.01)}$ & $0.81 \text{\tiny (0.01)}$ & $\boldsymbol{0.90} \text{\tiny (0.00)}$ & $\boldsymbol{0.97} \text{\tiny (0.00)}$ & $0.91 \text{\tiny (0.01)}$ & $0.82 \text{\tiny (0.02)}$ & $0.92 \text{\tiny (0.01)}$ & $0.81 \text{\tiny (0.01)}$ & $0.73 \text{\tiny (0.02)}$ & $\boldsymbol{0.98} \text{\tiny (0.00)}$ & $0.89 \text{\tiny (0.01)}$ & $0.81 \text{\tiny (0.02)}$  \\
         \multicolumn{3}{c}{Random} & $0.53 \text{\tiny (0.02)}$  & $0.53 \text{\tiny (0.02)}$ & $0.51 \text{\tiny (0.01)}$ & $0.51 \text{\tiny (0.01)}$ & $0.49 \text{\tiny (0.02)}$ & $0.49 \text{\tiny (0.02)}$ & $0.51 \text{\tiny (0.01)}$ & $0.51 \text{\tiny (0.01)}$ & $0.50 \text{\tiny (0.01)}$ & $0.51 \text{\tiny (0.01)}$ & $0.51 \text{\tiny (0.01)}$ & $0.38 \text{\tiny (0.02)}$ & $0.51 \text{\tiny (0.01)}$ & $0.51 \text{\tiny (0.01)}$  \\
        \bottomrule
    \end{tabular}
\end{table*}

\subsubsection{Training Setting}
The considered learning models are trained for up to 100 epochs with early stopping based on validation accuracy. 
The training and testing samples are randomly selected from the datasets, with the training size selected from $\{200,1000\}$. For datasets from the House of Graphs, the testing size is $500$ (resp., $200$) for Ham, Planar, and Claw (resp., Tree), while for synthetic datasets, the test set size is fixed at $500$ for all tasks. Each training set is further split into $80\%$ for training and $20\%$ for validation. 
As for the adversarial pretraining step (i.e., Eq.~\ref{eq: pre_train}), a separate pretraining set is randomly sampled from the remaining data: $2000$ samples for Ham and Planar, $1000$ for Claw, and $200$ for Tree, using datasets from House of Graphs; and $2000$ samples per task for the synthetic datasets.
Each method is repeated five times with different random seeds, and we report the average F1 score and standard deviations. Detailed training settings can be found in \Cref{apd:training_settings}, including hyperparameter settings, learning rates, optimizers, and configuration of computing platforms.

\subsection{Experimental Results Analysis}
In this section, we analyze the comparative results and also present studies to verify that VSAL indeed operates as expected.
The main results are summarized in Table~\ref{tab:F1 Hamiltonian Cycle Problem_large_huge}, where all VSAL variants use circular initial layouts.
The complete results, including those of VSAL with different initial layouts, as well as additional analysis, are provided in Appendix~\ref{apd:addtional_comparison}.

\begin{figure}[t!]
\centering
\subfloat{\includegraphics[width=0.115\textwidth]{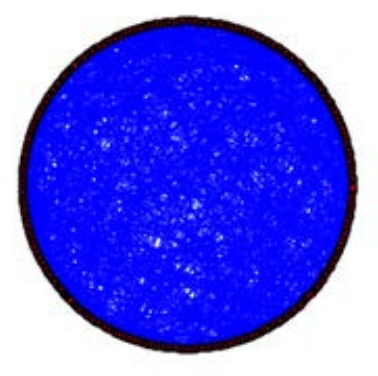}}\hspace{0mm}
\subfloat{\includegraphics[width=0.115\textwidth]{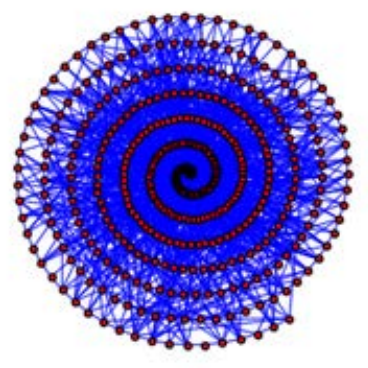}}\hspace{0mm}
\subfloat{\includegraphics[width=0.115\textwidth]{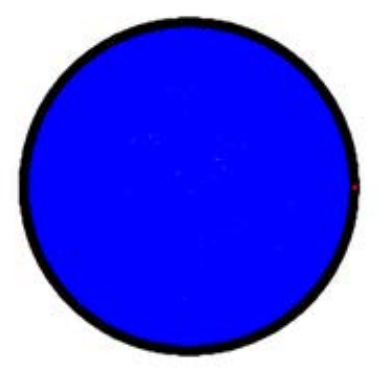}}\hspace{0mm}
\subfloat{\includegraphics[width=0.115\textwidth]{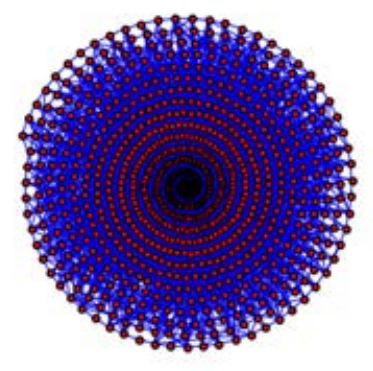}}
\setcounter{subfigure}{0} 
\renewcommand{\thesubfigure}{\alph{subfigure}} 

\subfloat[large circular]{\label{fig:vn-large-layout}\includegraphics[width=0.115\textwidth]{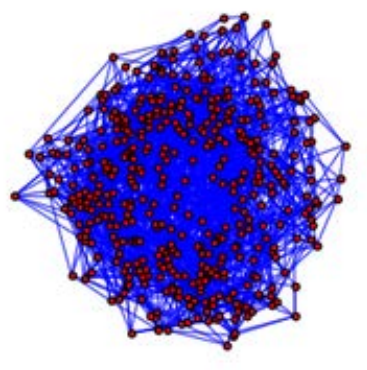}}\hspace{0mm}
\subfloat[large spiral]{\label{fig:vsgl-large-layout}\includegraphics[width=0.115\textwidth]{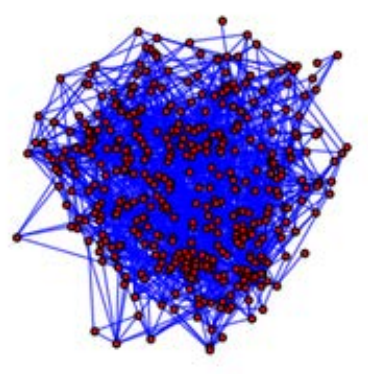}}\hspace{0mm}
\subfloat[huge circular]{\label{fig:vn-huge-layout}\includegraphics[width=0.115\textwidth]{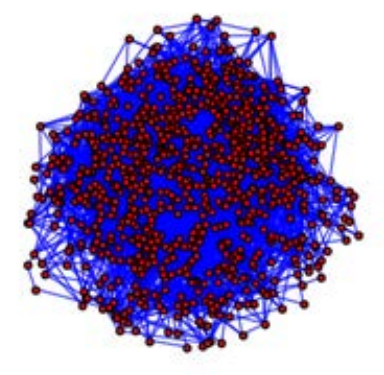}}\hspace{0mm}
\subfloat[huge spiral]{\label{fig:vsgl-huge-layout}\includegraphics[width=0.115\textwidth]{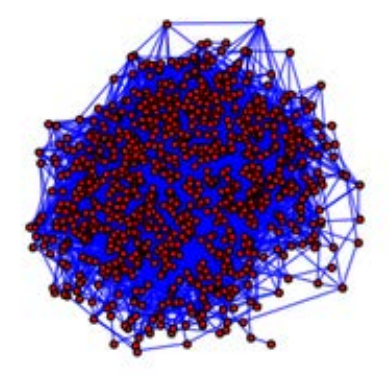}}

\Description{This figure shows the heavy overlappings in the large and huge graphs.}
\caption{Each pair illustrates graph layouts produced by VN-Solver (top) and VSAL (bottom) on the Ham. (a) and (b) show large graph layouts with circular and spiral layouts. 
(c) and (d) present the corresponding layouts for huge graphs. 
}
\label{fig:vn_solver_vsgl_layouts}
\end{figure}

\begin{figure*}[t!]
\centering
\subfloat[Generated layouts with circular initial layout on small graph]{\label{fig:Circular_small}\includegraphics[width=0.10\textwidth]{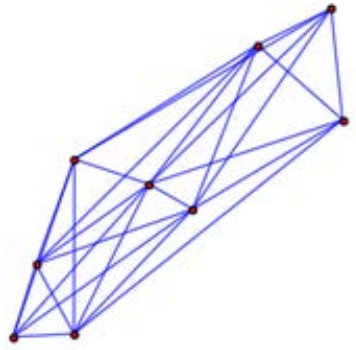}\hspace{3mm}
\includegraphics[width=0.10\textwidth]{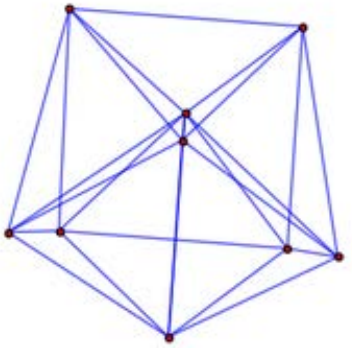}\hspace{3mm}
\includegraphics[width=0.10\textwidth]{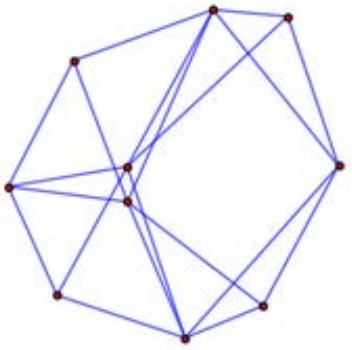}\hspace{3mm}
\includegraphics[width=0.10\textwidth]{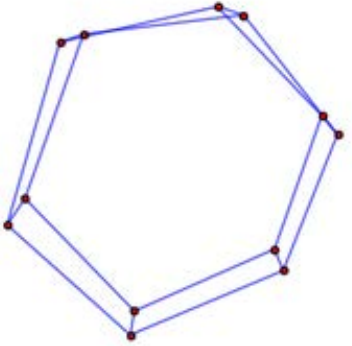}} 
\hspace{10mm}
\subfloat[Generated layouts with circular initial layout on medium graph]{\label{fig:Circular_large}\includegraphics[width=0.10\textwidth]{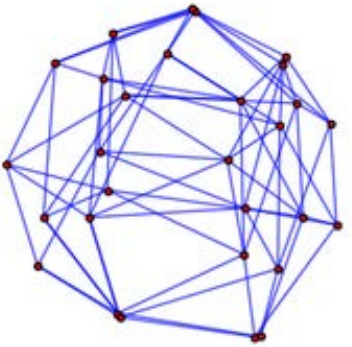}\hspace{3mm}
\includegraphics[width=0.10\textwidth]{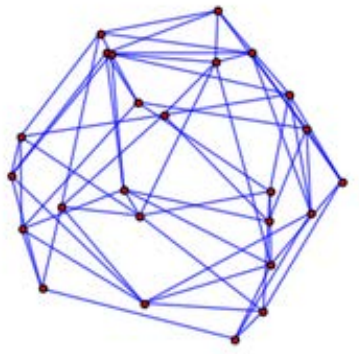}\hspace{3mm}
\includegraphics[width=0.10\textwidth]{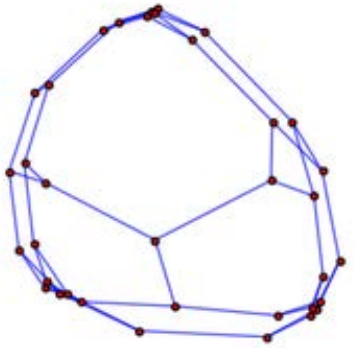}\hspace{3mm}
\includegraphics[width=0.10\textwidth]{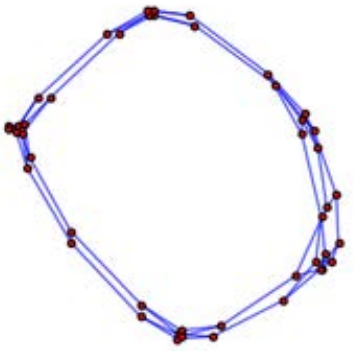}}  

\subfloat[Generated layouts with circular initial layout on large graph]{\label{fig: circular_401-500}\includegraphics[width=0.10\textwidth]{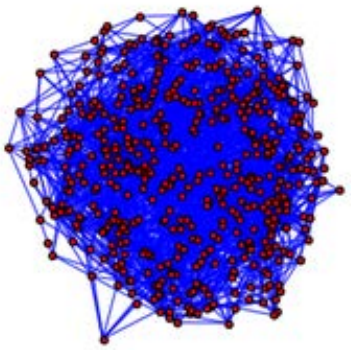}\hspace{3mm}
\includegraphics[width=0.10\textwidth]{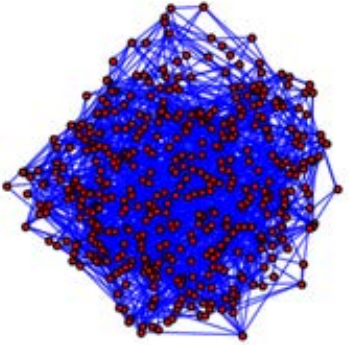}\hspace{3mm}
\includegraphics[width=0.10\textwidth]{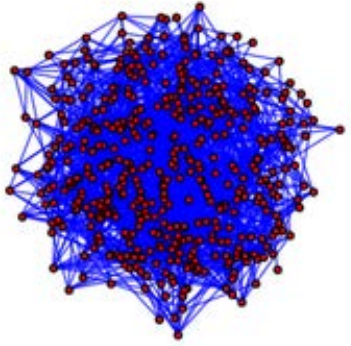}\hspace{3mm}
\includegraphics[width=0.10\textwidth]{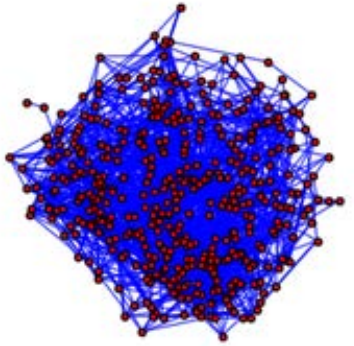}}\hspace{10mm}
\subfloat[Generated layouts with spiral initial layout on medium graph]{\label{fig:spiral_large}\includegraphics[width=0.10\textwidth]{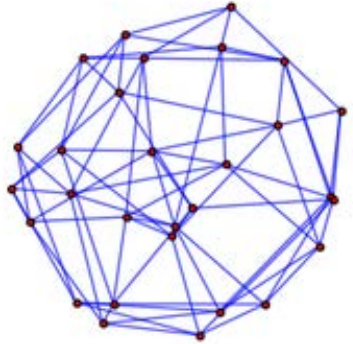}\hspace{3mm}
\includegraphics[width=0.10\textwidth]{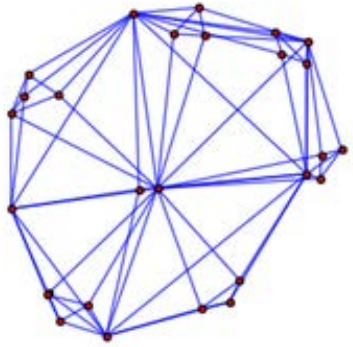}\hspace{3mm}
\includegraphics[width=0.10\textwidth]{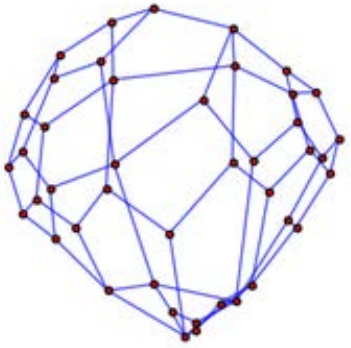}\hspace{3mm}
\includegraphics[width=0.10\textwidth]{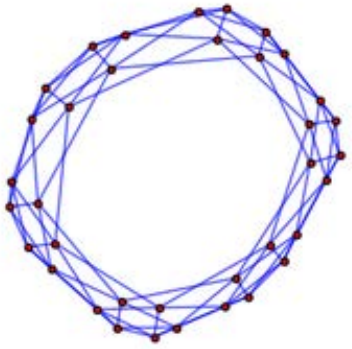}}\hspace{8mm}

\subfloat[Generated layouts with shell initial layout on medium graph]{\label{fig:shell_large}\includegraphics[width=0.10\textwidth]{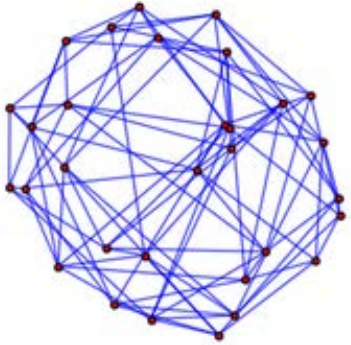}\hspace{3mm}
\includegraphics[width=0.10\textwidth]{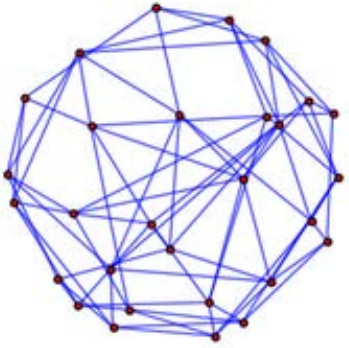}\hspace{3mm}
\includegraphics[width=0.10\textwidth]{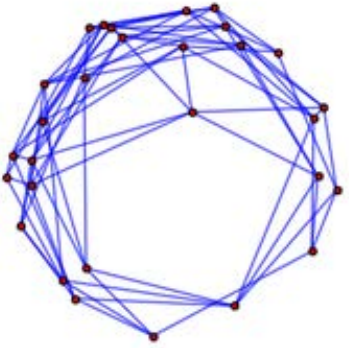}\hspace{3mm}
\includegraphics[width=0.10\textwidth]{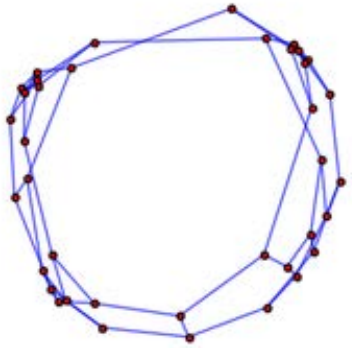}}\hspace{8mm}
\subfloat[Generated layouts with uniform initial layout on medium graph]{\label{fig: uniform_large}\includegraphics[width=0.10\textwidth]{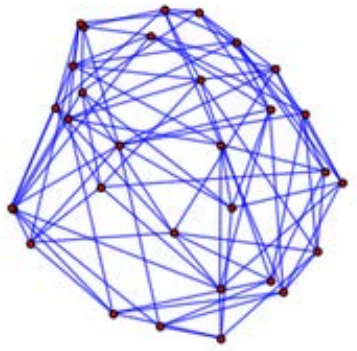}\hspace{3mm}
\includegraphics[width=0.10\textwidth]{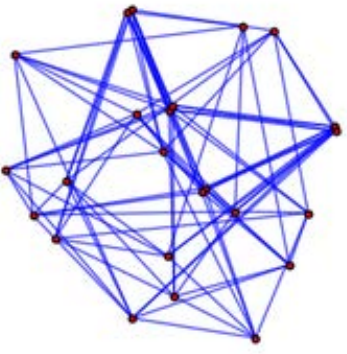}\hspace{3mm}
\includegraphics[width=0.10\textwidth]{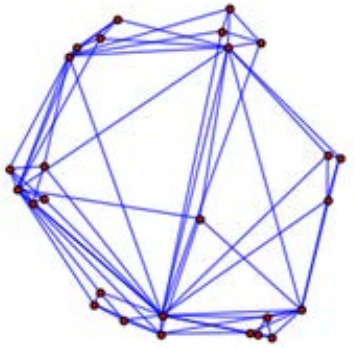}\hspace{3mm}
\includegraphics[width=0.10\textwidth]{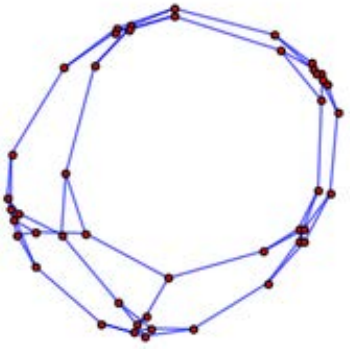}}

\Description{This figure depicts the dynamic changes of the generated graph layouts during the training.}
\caption{\small Examples of graph layouts generated during the training of VSAL on the Hamiltonian cycle problem at a resolution of $224\times 224$. Each group shows four layouts generated after, respectively, 0, 5, 10, and 15 epochs, from left to right. 
}

\label{fig:visualization_generated_graphs}
\end{figure*}

\textbf{VSAL outperforms VN-Solver.} 
VSAL consistently achieves better results over VN-Solver, which relies on fixed layouts, by an evident margin, demonstrating the effectiveness of adaptive layout generation; for example, VSAL-V with Graphormer improves the F1 score from $0.76$ to $0.92$ on the Ham with huge graphs
and from $0.83$ to $0.98$ on the Tree with small graphs.
Notably, VSAL still maintains strong performance on large and huge graphs. 
By comparison, the results of VN-Solver heavily rely on the choice of fixed layouts; for instance, on the Ham with large and huge graphs, circular layouts approach random guessing, whereas spiral layouts yield effective results. 
This reveals that different fixed layouts may only work under specific scenarios. As illustrated in Fig.~\ref{fig:vn_solver_vsgl_layouts}, fixed circular layouts often lead to severe edge overlaps in larger graphs, obscuring structural features, while spiral layouts can retain partial useful features. 
In contrast, the learned layouts of VSAL exhibit reduced edge overlaps and better preserve key features for detection.

\begin{table}[t] 
\renewcommand{\arraystretch}{1}
\centering 
\caption{F1 scores of different initial layouts on the Ham.} 
\label{tab:layout} 
\begin{tabular}{@{}l@{\hskip 12pt}c@{\hskip 8pt}c@{\hskip 12pt}c@{\hskip 8pt}c@{}} 
\toprule \multirow{2}{*}{\makecell{Train size \\ 1000}} &  \multicolumn{2}{c}{VSAL-V-224}&  \multicolumn{2}{c}{VN-Solver}\\
\cmidrule(lr){2-3}\cmidrule(lr){4-5}
 & Small  & Medium & Small  & Medium \\
\midrule 
Circular & $0.90 \text{\tiny (0.01)}$ & $0.98 \text{\tiny (0.00)}$ & $0.83  \text{\tiny (0.03)}$ & $0.95 \text{\tiny (0.01) }$\\
Spiral & $0.90 \text{\tiny (0.01)}$ & $0.98 \text{\tiny (0.01)}$ & $0.80 \text{\tiny (0.02)}$ & $0.95 \text{\tiny (0.01)}$\\
Shell  & $0.89  \text{\tiny (0.01)}$ & $0.97 \text{\tiny (0.00)}$ & $0.74 \text{\tiny (0.05)}$ & $0.94 \text{\tiny (0.02)}$\\
Uniform  & $0.87 \text{\tiny (0.01)}$ & $0.96 \text{\tiny (0.01)}$ & $0.58 \text{\tiny (0.01)}$ & $0.65 \text{\tiny (0.00)}$\\
\bottomrule 
\end{tabular} 
\end{table}

\textbf{VSAL maintains robustness across various initial layouts.} Table~\ref{tab:layout} presents the results of DenseGCN-based VSAL and VN-Solver using different initial layouts, i.e., circular (Fig.~\ref{fig:Circular_large}), spiral (Fig.~\ref{fig:spiral_large}), shell (Fig.~\ref{fig:shell_large}), and uniform (Fig.~\ref{fig: uniform_large}), on the Ham; complete results for other tasks are provided in Table~\ref{tab:different_initial_layout_ham} (Appendix~\ref{apd:addtional_comparison}). As expected, structured initial layouts with clear spatial organization provide more informative visual representations for detection than uniform layouts. Interestingly, from Table~\ref{tab:layout}, VSAL maintains high F1 scores across different initial layouts including uniform ones, which proves its ability to learn to generate effective layouts even from less principled initial layouts. For instance,  VSAL-V achieves F1 scores ranging from $0.96$ to $0.98$ across different initial layouts on medium graphs. In contrast, the performance of VN-Solver drops sharply with less structured layouts; for example, from $0.95$ (circular and spiral) to $0.65$ (uniform) on medium graphs. 

\begin{figure}
    \centering   \subfloat[Large layouts of VSAL]{\includegraphics[width=0.11\textwidth]{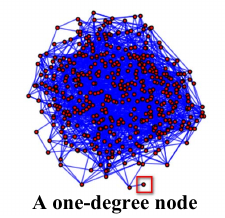}\hspace{2mm}\includegraphics[width=0.11\textwidth]{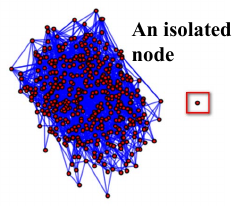}} \subfloat[Huge layouts of VSAL]{\includegraphics[width=0.11\textwidth]{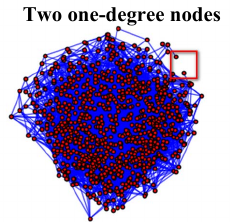}\hspace{2mm}\includegraphics[width=0.11\textwidth]{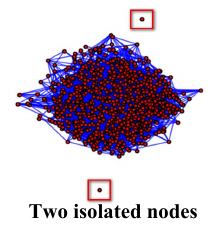}}
    \Description{This figure illustrates the visual cues that can be captured by VSAL, like one-degree nodes and isolated nodes.}
    \caption{VSAL-generated layouts on the Ham highlight structural features critical for prediction. 
    In these layouts, one-degree nodes are distinctly positioned and isolated nodes are clearly separated,
    thus providing visual cues that help the model better infer the absence of Hamiltonian cycles.}
    \label{fig:VSGL-extracted-features}
\end{figure}

\textbf{VSAL-generated layouts enhance clarity and capture useful features.} Fig.~\ref{fig:visualization_generated_graphs} illustrates the dynamic changes of generated layouts throughout training on the Ham. For small and medium graphs, the layouts exhibit increasing structural clarity with fewer edge crossings, 
offering more useful features
that facilitate the detection of the Hamiltonian cycle. In the case of large and huge graphs, the relatively small image resolution constrains structural clarity (Fig.~\ref{fig: circular_401-500});
nevertheless, the generated layouts can still emphasize key visual features. For example, as shown in Fig.~\ref{fig:VSGL-extracted-features}, the generated layouts on the Ham effectively capture visual cues such as isolated nodes and one-degree nodes---both critical for the Hamiltonian cycle problem. Such observations demonstrate that VSAL effectively learns to generate graph layouts in an expected manner and underscore the significance of incorporating classifier feedback into the training of the generator.

\textbf{VSAL with Graphormer encoder outperforms VSAL with DenseGCN.} We assess VSAL using two representative graph encoders: Graphormer and DenseGCN. 
As reported in Table~\ref{tab:F1 Hamiltonian Cycle Problem_large_huge}, the Graphormer-based variant consistently achieves higher F1 scores across multiple tasks. For instance, on the Ham with small graphs, VSAL-V with Graphormer reaches an F1 score of $0.93$, compared to $0.90$ with DenseGCN. This improvement aligns with our expectation, as Graphormer can capture long-range dependencies and global structural features effectively, offering stronger expressive power than DenseGCN. This enables VSAL to generate more informative layouts and achieve higher classification accuracy.

\textbf{VSAL-V outperforms VSAL-R and VSAL-E at the same resolution.} While VSAL-V, VSAL-R, and VSAL-E all achieve strong results, VSAL-V shows a slight advantage, particularly 
on small graphs (see Table~\ref{tab:F1 Hamiltonian Cycle Problem_large_huge}). 
For instance, on the Planar with medium graphs and $224\times 224$ resolution, all three DenseGCN-based variants achieve the same F1 scores of $0.97$. However, when the graph size is changed to small, VSAL-V outperforms others, achieving an F1 score of $0.89$ compared to $0.86$ for VSAL-R and $0.87$ for VSAL-E.
The results echo the fact that ViT is in general more effective in capturing visual patterns than ResNet-50 and EfficientNet.

\textbf{Higher resolution achieves better performance.} We evaluate VSAL at multiple image resolutions and observe that resolution influences detection accuracy, especially for large and huge graphs. As shown in Table~\ref{tab:F1 Hamiltonian Cycle Problem_large_huge}, resolution has little impact on small and medium graphs, where the lower density of nodes and edges enables visual clarity even at limited resolutions. In contrast, large and huge graphs exhibit increased visual clutter at low resolutions, and higher resolutions preserve fine-grained structural details that enhance detection.
For instance, on the Ham with large graphs,  DenseGCN-based VSAL-E achieves an F1 score of $0.94$ at $528\times 528$ resolution, compared to $0.92$ at $380 \times 380$ and $0.89$ at $224 \times 224$.  

\begin{table}[t!] 
\renewcommand{\arraystretch}{1}
\centering 
\caption{Ablation results (F1 scores) of Graphormer-based VSAL-V-224. 
``S'' and ``M'' denotes small and medium graphs. Exp-A: w/o Pretraining, Exp-B: w/o Refinement, Exp-C: w/o Gaussian Smoothing, Exp-D: w/o Visualization Module.} 
\label{tab:ablation_small} 
\begin{tabular}{@{}l@{\hskip 4 pt}cccccc@{}} 
\toprule 
& & VSAL-V & Exp-A  & Exp-B & Exp-C & Exp-D \\
\midrule 
\multirow{2}{*}{Ham}& S  & $0.93 \text{\tiny (0.01)}$ & $0.88 \text{\tiny (0.01)}$ & $0.89 \text{\tiny (0.01)}$ & $0.91 \text{\tiny (0.01)}$  & $0.90 \text{\tiny (0.00)}$  \\ 
& M  & $0.98 \text{\tiny (0.00)}$ & $0.94 \text{\tiny (0.01)}$ & $0.95 \text{\tiny (0.01)}$ & $0.97 \text{\tiny (0.01)}$   & $0.96 \text{\tiny (0.00)}$ \\ 
\multirow{2}{*}{Planar}& S  & $0.92 \text{\tiny (0.01)} $ & $0.87 \text{\tiny (0.02)}$ & $0.88 \text{\tiny (0.01)}$ & $0.90 \text{\tiny (0.01)}$ & $0.89 \text{\tiny (0.01)}$ \\ 
& M  & $0.98 \text{\tiny (0.00)}$ & $0.95 \text{\tiny (0.01)}$ & $0.94 \text{\tiny (0.01)}$ & $0.96 \text{\tiny (0.01)}$ & $0.95 \text{\tiny (0.01)}$ \\ 
Claw & S & $0.95  \text{\tiny (0.01) }$  & $0.90 \text{\tiny (0.02)}$ & $0.91 \text{\tiny (0.01)}$ & $0.93 \text{\tiny (0.01)}$ & $0.92 \text{\tiny (0.00)}$ \\
Tree& S & $0.98 \text{\tiny (0.01)}$ & $0.93 \text{\tiny (0.02)}$ & $0.93 \text{\tiny (0.01)}$ & $0.95 \text{\tiny (0.01)}$ & $0.94 \text{\tiny (0.01)}$ \\
\bottomrule 
\end{tabular} 
\end{table}

\textbf{The impact of critical modules in VSAL.}
\label{subsec:Ablation}
We conduct ablation studies through the following experiments using Graphormer-based VSAL-V: (Exp-A) VSAL-V without pretraining (Eq.~\ref{eq: pre_train}), (Exp-B) VSAL-V without the refinement module (Eq.~\ref{eq: refine}), (Exp-C) VSAL-V without Gaussian smoothing (Eq.~\ref{eq:gaussian_smoothing}), and (Exp-D) VSAL-V without the visualization module (Sec.~\ref{sec:visualization}). 
The first three cases ablate individual modules, 
whereas Exp-D removes the entire visualization module and instead employs a customized Graphormer classifier to directly process the generated layouts for label prediction.
The results in Table~\ref{tab:ablation_small} indicate that all modules contribute to improved detection performance. Notably, the degradation in Exp-D confirms the necessity of the visualization module in VSAL.
Additional results on large and huge graphs are provided in Appendix~\ref{apd:addtional_comparison}.

\textbf{VSAL outperforms matrix-based methods.} 
As shown in Table~\ref{tab:F1 Hamiltonian Cycle Problem_large_huge},  
VSAL surpasses state-of-the-art matrix-based methods, including Graphormer, Graphormer-GD, EquiformerV2, and GraphsGPT on most tasks, particularly on the large and huge graphs, and achieves comparable performance in the remaining cases. For example, on the Claw with large graphs, DenseGCN-based VSAL-V achieves an F1 score of $0.93$, outperforming Graphormer ($0.80$), Graphormer-GD ($0.83$), EquiformerV2 ($0.77$), and GraphsGPT ($0.81$). 
These results confirm again the effectiveness of vision-based approaches, which not only achieve promising accuracy but also offer improved interpretability through explicit layout visualization.

\textbf{VSAL achieves higher efficiency in terms of inference time and memory usage.} We compare the inference time of DenseGCN-based VSAL with several exact methods on different tasks, as detailed in Appendix~\ref{apd:addtional_comparison}. The results in 
Table~\ref{tab:runtime} (Appendix~\ref{apd:addtional_comparison}) 
demonstrate that VSAL can achieve extreme efficiency without sacrificing much efficacy; for instance, VSAL-V processes a single huge graph in just 0.015 seconds, whereas the Held-Karp algorithm  \cite{held1962dynamic} takes over 3600 seconds on the Ham. Additionally, we compare GPU memory usage between DenseGCN-based VSAL and Graphormer, as shown in Table~\ref{tab:memory_usgae} (Appendix~\ref{apd:addtional_comparison}). With graph size increasing, VSAL maintains significantly lower memory usage; for instance, 2965.8 MB on huge graphs compared to 39785.8 MB of Graphormer. This is because the memory usage of VSAL depends on image resolution (set to $224 \times 224$ here, but adjustable), while Graphormer scales with adjacency matrix size ($1000 \times 1000$ for huge graphs). Although image representations may abstract some details, their resolution can be tuned to balance efficiency and accuracy, thus enabling VSAL to process much larger graphs under limited computational resources.

\section{Future Discussion}
\label{sec:coclusion}
In this paper, we propose VSAL, a new vision-based framework that can generate adaptive layouts for various graph property detection tasks. We conclude by offering several interesting directions:

\textbf{Task-specific vision-based methods.} While VSAL achieves better results than several advanced general-purpose matrix-based methods such as Graphormer and GraphsGPT,
it does not rule out the scenario that VSAL can be less effective than matrix-based methods tailored to specific tasks \cite{sun2023difusco}.
Future work could incorporate task-specific features to enhance detection, such as explicitly visualizing potential cycle structures for Hamiltonian cycles.

\textbf{Exploring visualization methods.} Technically, any mapping $\mathcal{G}\rightarrow\mathbb{R}^{H\times W \times 3}$ can serve as the graph visualization module, leaving great room to explore more effective visualization strategies. 
A direction is to investigate how visual attributes, such as node size, edge thickness, and color schemes, affect detection performance.  
For example, enlarging or recoloring isolated nodes  (as in Fig.~\ref{fig:VSGL-extracted-features}) may help Hamiltonian cycle detection.
Another direction is to design more efficient and fully differentiable visualization methods, such as differentiable rasterization, to better integrate layout generation and classification while reducing computational overhead.



\textbf{Advanced generative models and alternative representation methods.} 
Advanced generative models (e.g., diffusion models \cite{ho2020denoising}) may offer higher generative quality, and therefore, integrating such methods into VSAL may lead to better generalization performance. The challenge in doing so is that training these models results in substantial computational overhead. 
Finally, enhancing the generator with more effective representations (e.g., incorporating subgraph embeddings) may improve its expressiveness and enable the generation of more informative layouts.  

\begin{acks}
This project is supported in part by National Science Foundation under IIS-2144285 and IIS-2414308. 
\end{acks}

\clearpage

\bibliographystyle{ACM-Reference-Format}
\bibliography{sample-sigconf}

\clearpage

\appendix

\begin{center}
\textbf{\LARGE Appendix}
\end{center}

\section{Spring and Kamada-Kawai Algorithms}
\label{apd:layout_algorithm}
In this section, we describe the two layout algorithms employed in our framework: the Spring algorithm \cite{fruchterman1991graph} and the Kamada-Kawai algorithm \cite{kamada1989algorithm}. The pseudocode for the Spring layout algorithm is provided in Algorithm~\ref{alg:spring}, while the pseudocode for the Kamada-Kawai layout algorithm  is shown in Algorithm~\ref{alg:Kamada-Kawai}.



\begin{algorithm}
\caption{Spring Layout Algorithm}
\begin{algorithmic}[1]
\label{alg:spring}
\STATE \textbf{Input:} Graph $G = (V, E)$
\STATE \textbf{Output:} Node positions

\STATE Initialize positions of all nodes randomly in a 2D space
\STATE Set parameters:
\STATE \quad Area (size of the canvas)
\STATE \quad $k \gets \sqrt{\text{area} / |V|}$ (optimal pairwise distance)
\STATE \quad Temperature $T$ (controls maximum displacement )

\REPEAT
    \STATE \textbf{Calculate repulsive forces:}
    \FOR{each pair of nodes $(u, v) \in V$}
        \STATE $F_{\text{rep}}(u, v) \gets -C_{\text{rep}} \cdot \frac{k^2}{\text{distance}(u, v)}$ 
    \ENDFOR

    \STATE \textbf{Calculate attractive forces:}
    \FOR{each edge $(u, v) \in E$}
        \STATE $F_{\text{att}}(u, v) \gets -C_{\text{att}} \cdot \frac{\text{distance}(u, v)^2}{k}$
    \ENDFOR

    \STATE \textbf{Update node positions:}
    \FOR{each node $u \in V$}
        \STATE Compute net force $F(u) = \sum F_{\text{rep}} + \sum F_{\text{att}}$
        \STATE Update position: $\text{pos}(u) \gets \text{pos}(u) + \min(T, ||F(u)||) \cdot \operatorname{dir}(F(u))$
    \ENDFOR

    \STATE Reduce temperature: $T \gets T \cdot \text{cooling\_factor}$
\UNTIL{System stabilizes or reaches maximum iterations}

\RETURN Final positions of nodes
\end{algorithmic}
\end{algorithm}

\begin{algorithm}
\caption{Kamada-Kawai Layout Algorithm}
\begin{algorithmic}[1]
\label{alg:Kamada-Kawai}
\STATE \textbf{Input:} Graph $G = (V, E)$, shortest path distances $d_{ij}$ \\ for all pairs of nodes
\STATE \textbf{Output:} Node positions

\STATE Initialize positions of all nodes randomly in a 2D space

\STATE Define the energy function:
\STATE \quad $E = \sum_{(i, j)} k_{ij} \cdot (\| \text{pos}(i) - \text{pos}(j) \| - d_{ij})^2$, 
\STATE \quad where $k_{ij} = \text{constant} / d_{ij}^2$

\REPEAT
    \FOR{each node $u \in V$}
        \STATE Compute partial derivatives of $E$:
        \STATE $ \frac{\partial E}{\partial x_u} = \sum_{j \neq u} k_{uj} \cdot (\text{pos}_x(u) - \text{pos}_x(j)) \cdot (\|\text{pos}(u) - \text{pos}(j)\| - d_{uj}) / \|\text{pos}(u) - \text{pos}(j)\|$
        \STATE $ \frac{\partial E}{\partial y_u} = \sum_{j \neq u} k_{uj} \cdot (\text{pos}_y(u) - \text{pos}_y(j)) \cdot (\|\text{pos}(u) - \text{pos}(j)\| - d_{uj}) / \|\text{pos}(u) - \text{pos}(j)\|$

        \STATE Update positions using gradient descent:
        \STATE $\text{pos}_x(u) \gets \text{pos}_x(u) - \text{learning\_rate} \cdot \frac{\partial E}{\partial x_u}$
        \STATE $\text{pos}_y(u) \gets \text{pos}_y(u) - \text{learning\_rate} \cdot \frac{\partial E}{\partial y_u}$
    \ENDFOR
    \STATE Check convergence: Stop if changes in positions are \\ below a threshold
\UNTIL{Energy $E$ converges or reaches maximum iterations}

\RETURN Final positions of nodes
\end{algorithmic}
\end{algorithm}

\section{Detailed Experiment Information}
In this section, we provide detailed descriptions of experimental tasks, comprehensive settings for the experiments, and additional experimental results and analysis.

\subsection{Experimental Tasks}
\label{apd:tasks_definition}
To comprehensively evaluate our framework, we consider four well-known graph property detection tasks: Hamiltonian cycle problem \cite{gould2014recent}, planarity verification \cite{hopcroft1974efficient}, claw-free graph classification \cite{faudree1997claw}, and tree recognition \cite{shasha2002algorithmics}. The details of them are described below.

1) \textbf{Hamiltonian cycle problem}: This task determines whether a graph contains a cycle that visits each node exactly once before returning to the starting node. As an NP-hard problem, it is computationally challenging and widely used as a benchmark for combinatorial optimization over graphs, with applications in various areas such as routing, scheduling, and bioinformatics. 

2)\textbf{ Planarity verification}: This task determines whether a graph can be embedded in the plane without edge crossings. Planarity verification is essential in applications like circuit design, geographical mapping, and network visualization, as planar graphs reduce visual complexity and enhance interpretability. 

3) \textbf{Claw-free graph classification}: A graph is Claw-free if it does not contain a star with three
leaves, $K_{1,3}$, as an induced subgraph. Claw-free graphs exhibit unique structural properties that are useful in studying network stability and chemical bonding. 

4) \textbf{Tree recognition}: A tree is a connected and acyclic graph with $n$ vertices and $n-1$ edges, commonly used to represent hierarchical data and network structures. Tree structure recognition is crucial in fields such as spanning tree algorithms, taxonomy generation, and decision-making processes.


\subsection{Synthetic Dataset}
\label{apd:dataset}
This section describes the procedures used to generate synthetic graph datasets at two scales: \textit{large} graphs with 401-500 nodes, and \textit{huge} graphs with 901-1000 nodes. All graphs are undirected, unweighted, and stored as adjacency matrices. 

\subsubsection{Hamiltonian vs. Non-Hamiltonian Graphs}\

\textbf{Hamiltonian graphs:} Each graph is initialized as a simple cycle on $n$ nodes, guaranteeing the existence of at least one Hamiltonian cycle. To introduce structural diversity, extra edges are randomly added between non-consecutive node pairs $(i, j)$ with a fixed probability $p_{\text{extra}} = 0.005$, where $j \neq (i+1) \mod n$ and $(i, j) \neq (0, n-1)$.

\textbf{Non-Hamiltonian graphs:} Starting from the same $n$-node cycle, we randomly remove a fixed number of cycle edges, between 40 and 80 for large graphs, and between 200 and 250 for huge graphs, to break potential Hamiltonian cycles. Subsequently, extra edges are added using the same probability $p_{\text{extra}}=0.005$, excluding the previously removed and wraparound edges. 

\subsubsection{Planar vs. Non-Planar Graphs}\

\textbf{Planar graphs:} Each graph begins as a uniformly random spanning tree with $n$ nodes, ensuring connectivity and planarity. 
Additional edges are then incrementally added by randomly sampling node pairs and accepting only those that preserve planarity, as verified using the Boyer-Myrvold planarity test \cite{hopcroft1974efficient}. This process continues until the graph reaches approximately $\lceil 1.6n \rceil$ edges, resulting in near-sparse planar graphs.

\textbf{Non-planar graphs:} Starting from a sparse planar graph, we randomly embed a forbidden subgraph, either $K_5$ or $K_{3,3}$, by adding the edges required to form it among randomly selected nodes. To maintain the total edge count, an equal number of unrelated edges are removed. This guarantees non-planarity by introducing a subgraph that violates Kuratowski’s theorem.

\subsubsection{Claw-Free vs. Non-Claw-Free Graphs}\

\textbf{Claw-free graphs:} These graphs are constructed as line graphs of random trees on $n+1$ nodes, which are guaranteed to be claw-free by definition. To reach the target edge density ($\rho = 0.008$ for large graphs and $\rho = 0.005$ for huge graphs), we uniformly add edges from the set of non-edges, while ensuring that the total edge count does not exceed the maximum possible for the given $n$.

\textbf{Non-claw-free graphs:} These graphs are derived from the claw-free graphs by explicitly planting $K_{1,3}$ (claw) subgraphs. We insert a fixed number of claws (15 per large graph and 50 per huge graph) by selecting disjoint node sets and adding three edges per claw. To maintain the original edge count, an equal number of unrelated edges are randomly removed. 

\subsubsection{Tree vs. Non-Tree Graphs}\

\textbf{Tree graphs:} Each instance is produced as a uniformly sampled spanning tree on $n$ nodes, containing exactly $n-1$ edges.

\textbf{Non-tree graphs:} Each graph is constructed by modifying a random spanning tree through either adding or removing edges. For edge addition, extra edges are sampled uniformly from the set of non-edges, with the number of added edges selected uniformly between $\lceil 0.05(n - 1) \rceil$ and $\lceil 0.12(n - 1) \rceil$. For edge removal, a comparable proportion of edges is randomly deleted. This process ensures that the resulting graph either contains cycles or becomes disconnected, and is therefore not a tree.


\subsection{Training Settings}
\label{apd:training_settings}
In this section, we provide a detailed description of training settings for VSAL and other learning baselines. In particular, all experiments are conducted using 4 NVIDIA L40S GPUs.

\begin{itemize} 
\item \textbf{VSAL}: 
A detailed summary of all hyperparameter settings in VSAL can be found in Table~\ref{tab:hyperparameter}. Moreover, the empirical analysis for $\lambda_{c}$, $\lambda_{\text{gp}}$, $m$, $r$ and $\delta$ are provided in \Cref{apd:hyperparameter}. 
\item \textbf{VN-Solver}: The VN-Solver is trained using the Adam optimizer with a learning rate of 1e-3 and employs two fixed layouts, including circular and spiral layouts.
\item \textbf{Graphormer} \cite{ying2021transformers} and \textbf{Graphormer-GD} \cite{zhangrethinking}: 
They are trained using Adam optimizer with a learning rate of 1e-5. 

\item \textbf{EquiformerV2}: We follow the settings of \cite{liaoequiformerv2} and train it using the AdamW optimizer with a learning rate of 1e-4.

\item \textbf{GraphsGPT}: We follow the settings of  \cite{gao2024graph}, and it is trained using the AdamW optimizer with a learning rate of 1e-4.

\end{itemize}

\subsection{Analysis of Key Hyperparameters}
\label{apd:hyperparameter}
We analyze the impact of key hyperparameters by conducting experiments using DenseGCN-based VSAL-V on the Hamiltonian cycle problem with 1000 medium training graphs. The evaluated hyperparameters include the classification loss weight $\lambda_c$, gradient penalty coefficient $\lambda_{\text{gp}}$, the number of sampled noise vectors $m$, the node influence factor $r$, and the edge influence factor $\delta$. The results are summarized in Table~\ref{tab:hyperparameter_c}.
Based on these results and grid search, we set $\lambda_c = 30$, $\lambda_{\text{gp}} = 5$, $m=10$, $r=2$ and $\delta=1$ for all experiments.

\begin{table}[t] 
\renewcommand{\arraystretch}{1}
\centering 
\caption{Hyperparameter settings in VSAL.} 
\label{tab:hyperparameter} 
\begin{tabular}{@{}cc@{}} 
\toprule 
Hyperparameter & Value \\
\midrule
Learning rate of generator & $0.0001$ \\
Learning rate of discriminator & $0.0005$ \\
Learning rate of classifier & $0.00001$ \\
Classification loss weight $\lambda_{c}$ & $30$ \\
Gradient penalty coefficient $\lambda_{\text{gp}}$ & $5$ \\
The number of sampled noise vectors $m$ & $10$ \\
Hidden dimension of generator $d_g$ & $128$ \\
Hidden dimension of discriminator $d_s$ & $128$ \\
Dimension of latent noise space $d_{\mathcal{Z}}$ & $128$ \\
Visualization height and width $H \times W$ & $224, 380, 528$ \\
Control the degree of node influence $r$ & $2$ \\
Control the degree of edge influence $\delta$ & $1$ \\
Control the sharpness of approximation $\beta$ & $10$ \\
The radius of the Gaussian kernel $K$ & $5$ \\
The strength of the Gaussian smoothing $\sigma$ & $2$ \\
The radius of the circular initial layout $r_1$ & $1$ \\
The offset factor of the spiral initial layout $r_2$ & $0.2$ \\
The number of shells in the shell initial layout $S$ & $2$ \\
The radius of the first shell $r_{j,1}$& $0.5$ \\
The radius of the second shell $r_{j,2}$& $1$ \\
The bound of the uniform initial layout $b$& $1$ \\
			
\bottomrule 
\end{tabular} 
\end{table}

\begin{table}[t] 
\renewcommand{\arraystretch}{1}
\centering 
\caption{Hyperparameter analysis for $\lambda_c$, $\lambda_{\text{gp}}$, $m$, $r$ and $\delta$.} 
\label{tab:hyperparameter_c} 
\begin{tabular}{@{}c@{\hskip 2.5pt}c|c@{\hskip 2.5pt}c|c@{\hskip 2.5pt}c|c@{\hskip 2.5pt}c|c@{\hskip 2.5pt}c@{}}
\toprule $\lambda_c$ & F1 &$\lambda_{\text{gp}}$ & F1& $m$ & F1& $r$ & F1& $\delta$ & F1  \\ \midrule 
 $1$ & $0.96 \text{\tiny(0.01)}$ & $1$ & $0.97 \text{\tiny(0.01)}$& $5$& $0.96 \text{\tiny(0.00)}$ & $1$ & $0.95 \text{\tiny(0.01)}$ & $0.5$ & $0.96 \text{\tiny(0.01)}$ \\
 $10$ & $0.97 \text{\tiny(0.01)}$ & $5$  & $0.98 \text{\tiny(0.00)}$ & $8$ & $0.97 \text{\tiny(0.00)}$ & $2$ & $0.98 \text{\tiny(0.00)}$ & $1$ & $0.98 \text{\tiny(0.00)}$  \\
 $30$ & $0.98 \text{\tiny(0.00)}$ & $10$ & $0.97 \text{\tiny(0.00)}$ & $10$ & $0.98 \text{\tiny(0.00)}$ & $5$ & $0.91 \text{\tiny(0.01)}$ & $2$ & $0.94 \text{\tiny(0.01)}$ \\ 
 $50$ & $0.95 \text{\tiny(0.01)}$ & $20$ & $0.96 \text{\tiny(0.01)}$ & $15$ & $0.97 \text{\tiny(0.00)}$ & $10$ & $0.86 \text{\tiny(0.01)}$ & $5$ & $0.88 \text{\tiny(0.01)}$ \\
\bottomrule 
\end{tabular} 
\end{table}

\subsection{Additional results and Analysis}
\label{apd:addtional_comparison}

\textbf{VSAL is robust to different initial layouts.} Table~\ref{tab:different_initial_layout_ham} compares the performance of VSAL with DenseGCN against VN-Solver on four tasks using different initial layouts. VSAL consistently achieves high performance regardless of the initial layouts, while the results of VN-Solver vary significantly. This confirms the effectiveness and robustness of adaptive layouts again.



\begin{table*}[t] 
\renewcommand{\arraystretch}{1}
\centering 
\caption{
F1 scores for different initial layouts across four tasks, using a training size of $1000$ (and $200$ for the small tree dataset).} 
\label{tab:different_initial_layout_ham} 
\begin{tabular}{@{}l@{\hskip 12pt}c@{\hskip 3pt}c@{\hskip 3pt}c@{\hskip 3pt}c@{\hskip 10pt}c@{\hskip 3pt}c@{\hskip 3pt}c@{\hskip 3pt}c@{\hskip 10pt}c@{\hskip 3pt}c@{\hskip 3pt}c@{\hskip 10pt}c@{\hskip 3pt}c@{\hskip 3pt}c@{}} 
\toprule  
&  \multicolumn{8}{c}{Hamiltonian cycle problem}&  \multicolumn{6}{c}{Claw-free classification}\\
\cmidrule(lr){2-9}\cmidrule(lr){10-15}
&  \multicolumn{4}{c}{VSAL-V-224 (DenseGCN)}&  \multicolumn{4}{c}{VN-Solver}&  \multicolumn{3}{c}{VSAL-V-224 (DenseGCN)}&  \multicolumn{3}{c}{VN-Solver}\\
\cmidrule(lr){2-5}\cmidrule(lr){6-9}\cmidrule(lr){10-12}\cmidrule(lr){13-15}
 & Small  & Medium & Large  & Huge & Small  & Medium & Large  & Huge & Small  & Large  & Huge& Small   & Large  & Huge \\
\midrule 
Circular & $0.90 \text{\tiny(0.01)}$ & $0.98 \text{\tiny(0.00)} $ & $0.91 \text{\tiny(0.00)}$ & $0.90 \text{\tiny(0.01)}$ & $0.83 \text{\tiny(0.03)}$ & $0.95\text{\tiny(0.01)}$ & $0.57 \text{\tiny(0.03)}$ & $0.53 \text{\tiny(0.01)}$ & $0.93 \text{\tiny(0.01)}$ & $0.93 \text{\tiny(0.01)}$ & $0.92 \text{\tiny(0.01)}$ & $0.86 \text{\tiny(0.01)}$ & $0.60 \text{\tiny(0.01)}$ & $0.58 \text{\tiny(0.02)}$   \\
Spiral & $0.90 \text{\tiny(0.01)}$ & $0.98 \text{\tiny(0.01)}$ & $0.91 \text{\tiny(0.00)}$ & $0.90 \text{\tiny(0.01)}$ & $0.80 \text{\tiny(0.02)}$ & $0.95 \text{\tiny(0.01)}$ & $0.81 \text{\tiny(0.01)}$ & $0.76 \text{\tiny(0.01)}$ & $0.93 \text{\tiny(0.01)}$ & $0.93 \text{\tiny(0.01)}$ & $0.91 \text{\tiny(0.01)}$ & $0.85 \text{\tiny(0.01)}$ & $0.68 \text{\tiny(0.02)}$ & $0.62 \text{\tiny(0.02)}$ \\
Shell  & $0.89 \text{\tiny(0.01)}$ & $0.97 \text{\tiny(0.00)}$ & $0.91 \text{\tiny(0.01)}$ & $0.89 \text{\tiny(0.00)}$ & $0.74 \text{\tiny(0.05)}$ & $0.94 \text{\tiny(0.02)}$ & $0.61 \text{\tiny(0.02)} $ & $0.59 \text{\tiny(0.01)}$ & $0.93 \text{\tiny(0.01)}$ & $0.93 \text{\tiny(0.01)}$ & $0.90 \text{\tiny(0.01)}$ & $0.86 \text{\tiny(0.02)}$ & $0.61 \text{\tiny(0.01)}$ & $0.58 \text{\tiny(0.01)}$ \\
Uniform  & $0.87 \text{\tiny(0.01)}$ & $0.96 \text{\tiny(0.01)}$&  $0.89 \text{\tiny(0.01)}$ & $0.87 \text{\tiny(0.00)}$ & $0.58 \text{\tiny(0.01)}$ & $0.65 \text{\tiny(0.00)}$ & $0.53 \text{\tiny(0.02)}$ & $0.51 \text{\tiny(0.02)}$ & $0.90 \text{\tiny(0.01)}$ & $0.89 \text{\tiny(0.01)}$&  $0.87 \text{\tiny(0.01)}$ & $0.73 \text{\tiny(0.01)}$ & $0.56 \text{\tiny(0.02)}$ & $0.52 \text{\tiny(0.01)}$\\

\midrule[0.5pt]\midrule[0.5pt]

&  \multicolumn{8}{c}{Planarity verification}&  \multicolumn{6}{c}{Tree recognition }\\
\cmidrule(lr){2-9}\cmidrule(lr){10-15}
\midrule 
Circular & $0.89 \text{\tiny(0.01)}$ & $0.97 \text{\tiny(0.00)}$ & $0.92 \text{\tiny(0.01)}$ & $0.91 \text{\tiny(0.01)}$ & $0.85 \text{\tiny(0.02)}$ & $0.95 \text{\tiny(0.01)}$ & $0.51 \text{\tiny(0.02)}$ & $0.51 \text{\tiny(0.01)}$ & $0.96 \text{\tiny(0.01)}$ & $0.94 \text{\tiny(0.01)} $ & $0.92 \text{\tiny(0.00)}$ & $0.83 \text{\tiny(0.02)}$ & $0.76 \text{\tiny(0.01)}$ & $0.64 \text{\tiny(0.02)}$  \\
Spiral & $0.89 \text{\tiny(0.01)}$ & $0.96 \text{\tiny(0.00)}$ & $0.92 \text{\tiny(0.01)}$ & $0.91 \text{\tiny(0.01)}$ & $0.84 \text{\tiny(0.01)}$ & $0.94 \text{\tiny(0.02)}$ & $0.53 \text{\tiny(0.01)}$ & $0.51 \text{\tiny(0.02)}$& $0.95 \text{\tiny(0.01)}$ & $0.94 \text{\tiny(0.01)}$ & $0.92 \text{\tiny(0.01)}$ & $0.81 \text{\tiny(0.03)}$ & $0.77 \text{\tiny(0.02)}$ & $0.68 \text{\tiny(0.02)}$ \\
Shell  & $0.89 \text{\tiny(0.01)}$ & $0.97 \text{\tiny(0.01)}$ & $0.91 \text{\tiny(0.01)}$ & $0.89 \text{\tiny(0.01)}$ & $0.84 \text{\tiny(0.01)}$ & $0.96 \text{\tiny(0.01)}$ & $0.53 \text{\tiny(0.01)}$ & $0.51 \text{\tiny(0.01)}$ & $0.96 \text{\tiny(0.01)} $ & $0.93 \text{\tiny(0.01)}$ & $0.91 \text{\tiny(0.01)}$ & $0.76 \text{\tiny(0.04)}$ & $0.71 \text{\tiny(0.03)}$ & $0.63 \text{\tiny(0.02)}$ \\
Uniform  & $0.86 \text{\tiny(0.02)}$ & $0.93 \text{\tiny(0.01)}$&  $0.88 \text{\tiny(0.01)}$ & $0.85 \text{\tiny(0.01)}$ & $0.74 \text{\tiny(0.02)}$ & $0.85 \text{\tiny(0.01)}$ & $0.51 \text{\tiny(0.01)}$ & $0.47 \text{\tiny(0.02)}$ & $0.92 \text{\tiny(0.01)}$ & $0.90 \text{\tiny(0.01)}$&  $0.87 \text{\tiny(0.01)}$ & $0.72 \text{\tiny(0.03)}$ & $0.66 \text{\tiny(0.02)}$ & $0.55 \text{\tiny(0.02)}$ \\
\bottomrule 
\end{tabular} 
\end{table*}

\textbf{Inference time comparison.} To evaluate the efficiency of VSAL, we compare its inference time with several traditional exact algorithms used for graph property detection:
\begin{itemize}
    \item \textbf{Hamiltonian cycle problem:} Held-Karp algorithm \cite{held1962dynamic}, with exponential time complexity $\Theta(2^n n^2)$.
\item \textbf{Planarity verification:} Hopcroft-Tarjan algorithm \cite{hopcroft1974efficient}, with linear time complexity $O(n)$.
\item \textbf{Claw-free graph classification:} Brute-force induced subgraph check for $K_{1,3}$, with time complexity $O(n^4)$.
\item \textbf{Tree recognition:} Depth-First Search (DFS), with linear time complexity $O(n)$.
\end{itemize}
As shown in Table~\ref{tab:runtime}, VSAL significantly reduces computational overhead compared to algorithms with time complexity greater than $O(n)$, particularly on huge graphs. These results confirm that VSAL can achieve extreme efficiency without sacrificing much efficacy, highlighting again the advantages of learning-based methods for graph property detection.

\textbf{GPU memory usage comparison.} We compare the GPU memory usage of VSAL and Graphormer during training with a batch size of 1, and the results are presented in Table~\ref{tab:memory_usgae}. VSAL demonstrates relatively stable memory consumption, ranging from 1261.8 MB on small graphs to 2965.8 MB on huge graphs. In contrast, the GPU memory usage of Graphormer increases dramatically, from 945.8 MB on small graphs to 39785.8 MB on huge graphs, indicating poor scalability. This difference stems from their data representations: the memory usage of VSAL depends on fixed image resolution (set to $224 \times 224$ here, but adjustable), whereas Graphormer scales with the size of the adjacency matrix ($1000 \times 1000$ for huge graphs). Although image representations may abstract some details, their resolution can be tuned to balance efficiency and accuracy. Indeed, the graph visualizations still retain key structural features for graph property detection, thus allowing VSAL to process much larger graphs while maintaining comparable performance and significantly lower memory overhead.

\begin{table}[t!] 
\renewcommand{\arraystretch}{1} 
\centering \caption{Average running time on one graph (small and huge graph) for different tasks (in seconds).} 
\label{tab:runtime} 
\begin{tabular}{@{}lccccc@{}} 
\toprule Task &Method & Small & Huge\\
\midrule 
\multirow{2}{*}{Ham} &Held-Karp algorithm & $1.376$ & $>3600$ \\
&VSAL-V-224 & $0.008$ & $0.015$ \\ 
\multirow{2}{*}{Planar} & Hopcroft-Tarjan algorithm & $0.00003$ & $0.0014$ \\
&VSAL-V-224 & $0.007$ & $0.012$ \\ 
\multirow{2}{*}{Claw}&Brute-force induced $K_{1,3}$
check & $0.016$ & $1361.258$ \\
&VSAL-V-224 & $0.007$ & $0.013$ \\ 
\multirow{2}{*}{Tree} &DFS on Tree & $0.000004$ & $0.0002$ \\
&VSAL-V-224 & $0.005$ & $0.009$ \\ 
\bottomrule 
\end{tabular} 
\end{table}

\begin{table}[t!] 
\renewcommand{\arraystretch}{1} 
\centering \caption{Average GPU memory usage during the training on the Hamiltonian cycle problem with a batch size of 1.} 
\label{tab:memory_usgae} 
\begin{tabular}{lcccc} 
\toprule Graph Size & VSAL-V-224 & Graphormer\\
\midrule Small &1261.8 MB   & 945.8 MB \\
Medium &1261.8 MB  & 1057.8 MB  \\ 
Large &2719.8 MB  &17521.8 MB  \\
Huge & 2965.8 MB  & 39785.8 MB  \\ 
\bottomrule 
\end{tabular} 
\end{table}

\begin{table}[t!] 
\renewcommand{\arraystretch}{1}
\centering 
\caption{Ablation results (F1 scores) of Graphormer-based VSAL-V.  ``L'' and ``H'' refer to large and huge graphs, respectively. Exp-A: w/o Pretraining, Exp-B: w/o Refinement, Exp-C: w/o Gaussian Smoothing, Exp-D: w/o Visualization Module.} 
\label{tab:ablation} 
\begin{tabular}{@{}l@{\hskip 5 pt}cccccc@{}} 
\toprule 
&  & VSAL-V & Exp-A  & Exp-B & Exp-C & Exp-D\\
\midrule 
\multirow{2}{*}{Ham}
& L  & $0.94 \text{\tiny(0.01)}$& $0.90 \text{\tiny(0.01)}$& $0.91 \text{\tiny(0.01)}$ & $0.92 \text{\tiny(0.01)}$ & $0.91 \text{\tiny(0.01)}$ \\ 
& H & $0.92 \text{\tiny(0.01)}$& $0.87 \text{\tiny(0.01)}$& $0.88 \text{\tiny(0.01)}$ & $0.89 \text{\tiny(0.00)}$ & $0.89 \text{\tiny(0.01)}$ \\ 
\multirow{2}{*}{Planar}
& L  & $0.94 \text{\tiny(0.01)}$& $0.89 \text{\tiny(0.01)}$& $0.91 \text{\tiny(0.01)}$ & $0.92 \text{\tiny(0.01)}$ & $0.91 \text{\tiny(0.01)}$ \\ 
& H  & $0.93 \text{\tiny(0.01)}$ & $0.87 \text{\tiny(0.01)}$ & $0.89 \text{\tiny(0.00)}$ & $0.90 \text{\tiny(0.01)}$ & $0.89 \text{\tiny(0.01)}$ \\ 
\multirow{2}{*}{Claw} 
& L  & $0.95 \text{\tiny(0.01)}$& $0.88 \text{\tiny(0.01)}$& $0.91 \text{\tiny(0.01)}$ & $0.93 \text{\tiny(0.01)}$ & $0.90 \text{\tiny(0.01)}$ \\ 
& H & $0.94 \text{\tiny(0.01)}$& $0.86 \text{\tiny(0.01)}$& $0.89 \text{\tiny(0.01)}$ & $0.91 \text{\tiny(0.01)}$ & $0.89 \text{\tiny(0.01)}$ \\ 
\multirow{2}{*}{Tree}
& L & $0.96 \text{\tiny(0.01)}$& $0.90 \text{\tiny(0.01)}$& $0.92 \text{\tiny(0.01)}$ & $0.94 \text{\tiny(0.00)}$ & $0.93 \text{\tiny(0.01)}$ \\ 
& H  & $0.94 \text{\tiny(0.01)}$& $0.87 \text{\tiny(0.01)}$& $0.90 \text{\tiny(0.01)}$ & $0.91 \text{\tiny(0.01)}$ & $0.90 \text{\tiny(0.01)}$ \\ 
\bottomrule 
\end{tabular} 
\end{table}

\textbf{Ablation study.}
The complementary results of ablation study on large and huge graphs are summarized in Table ~\ref{tab:ablation}. This further confirm that all these modules are helpful in improving graph property detection performance.

\end{document}